\pdfoutput=1
\documentclass[11pt]{article}

\usepackage[final]{acl}

\usepackage{times}
\usepackage{latexsym}
\usepackage[T1]{fontenc}
\usepackage[utf8]{inputenc}
\usepackage{microtype}
\usepackage{inconsolata}
\usepackage{graphicx}
\usepackage{amsmath}
\usepackage{amssymb}
\usepackage{amsthm}
\usepackage{booktabs}
\usepackage{multirow}
\usepackage[ruled,vlined,linesnumbered]{algorithm2e}
\usepackage{float}
\usepackage{enumitem}
\usepackage{pifont}

\makeatletter
\def\AutoCRAT@H{H}
\newif\ifAutoCRAT@algH
\renewenvironment{algorithm}[1][htbp]{%
  \def\AutoCRAT@algplacement{#1}%
  \setboolean{algocf@algostar}{false}%
  \setboolean{algocf@procenvironment}{false}%
  \ifx\AutoCRAT@algplacement\AutoCRAT@H
    \AutoCRAT@algHtrue
    \@algocf@init%
    \@algocf@init@caption%
    \setboolean{algocf@algoH}{true}\begin{algocf@Here}%
      \ifthenelse{\boolean{algocf@customruledwidth}}{\relax}{\setlength{\algocf@ruledwidth}{\linewidth}}%
      \let\algocf@oldeverypar=\everypar%
      \algocf@seteverypar%
    \@algocf@start%
    \@ResetCounterIfNeeded%
    \algocf@linesnumbered\ignorespaces%
  \else
    \AutoCRAT@algHfalse
    \begin{algocf@algorithm}[#1]\ignorespaces%
  \fi
}{%
  \ifAutoCRAT@algH
    \@algocf@finish%
    \@algocf@term@caption%
    \let\everypar=\algocf@oldeverypar%
    \end{algocf@Here}\par%
    \@algocf@term\ignorespacesafterend%
  \else
    \end{algocf@algorithm}\ignorespacesafterend%
  \fi
}
\makeatother

\definecolor{GainRed}{RGB}{214,39,40}
\definecolor{GainGreen}{RGB}{34,139,34}
\SetKwInput{KwInput}{Input}
\SetKwInput{KwOutput}{Output}
\newcommand{\EqMark}[1]{\hfill\mbox{\color{blue}$\triangleright$~Eq.~\eqref{#1}}}
\newcommand{\AlgComment}[1]{\textcolor{blue}{// #1}}
\newcommand{\AccGain}[1]{$_{\color{GainRed}\uparrow #1}$}
\newcommand{\TokGain}[1]{$_{\color{GainRed}\downarrow #1\%}$}
\newcommand{\AccDrop}[1]{$_{\color{GainGreen}\downarrow #1}$}
\newcommand{\TokOver}[1]{$_{\color{GainGreen}\uparrow #1\%}$}
\newcommand{\TokUnder}[1]{$_{\color{GainRed}\downarrow #1\%}$}

\newtheorem{proposition}{Proposition}

\title{AutoCRAT: Within-trajectory Joint Control of Stochasticity and Compute for LLM Reasoning}

\author{
  Hanjun Luo$^{1,2}$\thanks{\url{hl6266@nyu.edu}},
  Qiushi Liu$^{4}$,
  Jingya Zhang$^{1}$,
  Haihong Pang$^{1}$,
  Jiaheng Wen$^{5}$,
  Yifei Ma$^{1}$,\\
  \textbf{Yu Yao}$^{6}$,
  \textbf{Chengxi Zhang}$^{5}$,
  \textbf{Hanrong Zhang}$^{7}$,
  \textbf{Yankai Chen}$^{3}$\thanks{Corresponding author},
  \textbf{Hanan Salam}$^{2}$\\
  $^{1}$New York University,
  $^{2}$New York University Abu Dhabi,\\
  $^{3}$Mohamed bin Zayed University of Artificial Intelligence,\\
  $^{4}$University of Washington Seattle,
  $^{5}$Harvard University,\\
  $^{6}$Massachusetts Institute of Technology,
  $^{7}$University of Illinois Chicago
}

\begin{document}
\maketitle

\begin{abstract}

Large language models (LLMs) achieve strong reasoning performance, which depends critically on inference-time decisions. Yet these decisions are commonly handled by static, one-size-fits-all policies, limiting adaptation to diverse tasks and reasoning stages. Recent adaptive methods partially address this limitation, but they primarily adapt either decoding stochasticity (how the model explores) or reasoning compute (how long the model reasons) in isolation, leaving their interaction within a single reasoning trajectory unmodeled. To address this challenge, we shift toward a \textbf{within-trajectory joint control} view, and instantiate it in \textbf{\texttt{AutoCRAT}}, a decoder-side controller for frozen backbones. Using only signals available during decoding, \textbf{\texttt{AutoCRAT}} jointly adjusts \emph{sampling stochasticity} and \emph{reasoning budget} during generation. \textbf{\texttt{AutoCRAT}} operates over a discrete action space and updates control decisions only at semantic boundaries, improving stability while remaining responsive to the evolving reasoning process. Comprehensive evaluation across \textbf{6} benchmarks demonstrates that \textbf{\texttt{AutoCRAT}} \textbf{(I)} uses $13.8\sim 52.7\%$ fewer inference tokens on average than recommended static configurations, \textbf{(II)} surpasses recommended static and adaptive baselines by $1.5 \sim 4.5\%$ in relative accuracy, and \textbf{(III)} enjoys strong cross-backbone transferability.

\end{abstract}

\section{Introduction}
\label{sec:intro}

Large language models (LLMs) have demonstrated strong performance on complex reasoning tasks such as mathematics and code generation \cite{chen2025reasoningera}. In practice, their reasoning performance depends not only on the underlying capabilities but also heavily on inference-time configurations, including decoding stochasticity (e.g., temperature, top-p) and reasoning compute (e.g., reasoning depth and allocation of test-time compute), which are often manually pre-specified \cite{ji2025ttcsurvey}. Even when adjustments are available, they are primarily limited to per-request configurations fixed before generation begins, or rely on external human intervention \cite{zou2026llmbasedhumanagentcollaborationinteraction,li2026prefix}, without a mechanism to adapt as the optimal configurations evolve within a single trajectory. For instance, early stages may require broader exploration, while later stages require consolidation and answer commitment. Consequently, such fixed configurations are often ill-suited to the varying needs across tasks and reasoning stages \cite{parashar2025sys2bench,li2026safetyreproconfigurationconditionalrankinstability,fan2026movingtargetlongitudinalaudit}.

\begin{figure}[H]
\centering
\resizebox{0.99\columnwidth}{!}{\includegraphics{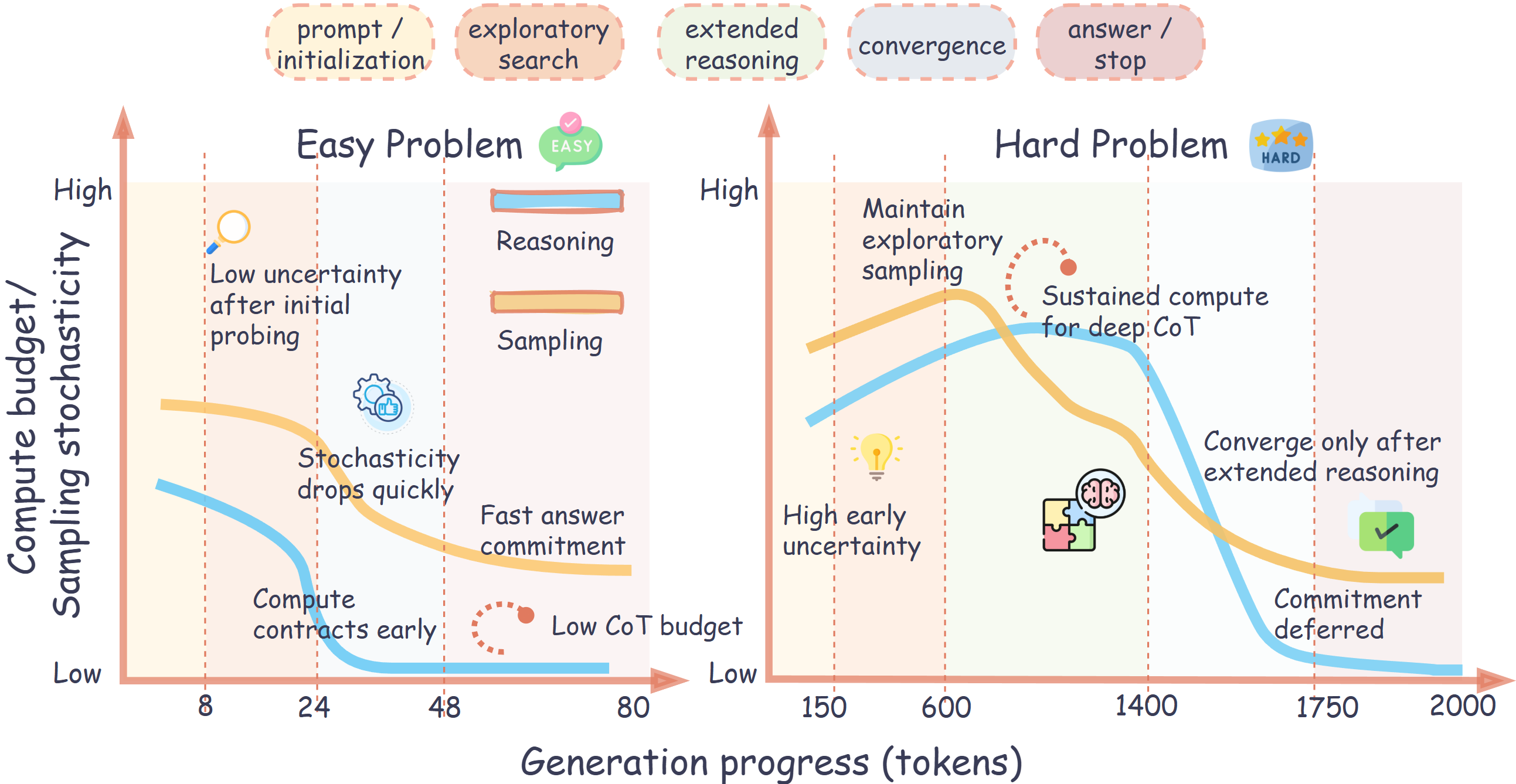}}
\caption{Existing methods adapt only part of the reasoning process, while the optimal workflow requires joint, within-trajectory coordination across stages.}
\label{fig:compare}
\end{figure}

Recent work has increasingly explored adaptive mechanisms for calibrating these factors during reasoning \cite{alomrani2025budgetsurvey}. One line of work adjusts decoding stochasticity through dynamic temperature scheduling or sampling adjustments \cite{zhu2024adaptivedecoding}; another line adjusts reasoning compute through adaptive thinking length, dynamic budget allocation, early stopping, or self-correction \cite{pu2025thoughtterminator,wu2025moreisless}. However, prior studies suggest that these two dimensions interact and should be coordinated jointly \cite{wu2024inferencescaling}, as demonstrated in Figure~\ref{fig:compare}. Yet existing methods still lack a unified within-trajectory control perspective that jointly models their interaction.  

To address this gap, we propose \textbf{\texttt{AutoCRAT}}, a decoder-side controller that performs joint within-trajectory coordination. Technically, we build \textbf{\texttt{AutoCRAT}} on a two-dimensional stochasticity-compute control view. It simultaneously regulates sampling stochasticity and reasoning budget as the generation unfolds. \textbf{\texttt{AutoCRAT}} operates solely on signals observable during decoding and its own control state, without accessing hidden states of backbones, which supports transferability across models and interfaces \cite{zhang2025capabilityboundaries,du2026searchtransferamortizedagentic}. To balance adaptivity and stability in this joint control setting, we introduce two designs: (i) control decisions are made over a discrete action space rather than continuous values, and (ii) are updated only at natural semantic boundaries (e.g., sentence endings, step markers), avoiding the instability of token-level switching while remaining more responsive than per-request configuration \cite{yu2025explainable}.

We conduct comprehensive evaluations on \textbf{6} widely adopted benchmarks, covering diverse use cases in code generation, mathematical reasoning, and challenging QA. Empirical results demonstrate that \textbf{\texttt{AutoCRAT}} is \textbf{(I) token-economical}, using $13.8\sim 52.7\%$ fewer inference tokens on average than recommended static configurations; \textbf{(II) high-performing}, surpassing recommended static and adaptive baselines by $1.5 \sim 4.5\%$ in relative accuracy; \textbf{(III) transferable} across LLM-backbones; \textbf{(IV) insightful}, demonstrating the significance of a unified view of LLM reasoning.

Our contributions are summarized as follows:

\begin{itemize}[leftmargin=1.6em,
    labelsep=0.5em,
    topsep=0.2em,
    partopsep=0pt,
    parsep=0pt,
    itemsep=0.15em]
    
    \item[\ding{182}] \textbf{Unified View.} We introduce a unified view of inference-time adaptation for LLM reasoning, arguing that stochasticity and compute should be coordinated within a trajectory rather than adjusted in isolation or only at the request level.

    \item[\ding{183}] \textbf{Practical Solution.} We propose \textbf{\texttt{AutoCRAT}}, a decoder-side controller for frozen backbones that jointly coordinates sampling stochasticity and reasoning budget. Its discrete action design and boundary-aware updates balance adaptivity, stability, and transferability.

    \item[\ding{184}] \textbf{Experimental Evaluation.} Extensive experiments on \textbf{6} benchmarks show that \textbf{\texttt{AutoCRAT}} achieves a substantially better accuracy--compute tradeoff than baselines, exhibits strong transferability, and provides insights into the dynamics of LLM reasoning control.
\end{itemize}

\section{Related Work}
\label{sec:related}

\textbf{Decoding Stochasticity Adaptation.} Decoding stochasticity is a central determinant of LLM behavior because sampling controls the tradeoff between exploration and exploitation during generation. Classical studies showed that decoding choices crucially affect degeneration and sample quality \cite{holtzman2020curious,shi2024decodingexam}. Later analyses further indicate that the effect of temperature on downstream reasoning is task-dependent rather than monotonic, making a single global configuration difficult to justify across diverse reasoning workloads \cite{renze2024temperature,salah2026temperatureextended}. Motivated by this sensitivity, recent work has moved beyond fixed per-request sampling rules toward adaptive stochasticity control. Representative directions include entropy-aware temperature adjustment \cite{zhang2024edt}, confidence- or entropy-based candidate-set adaptation \cite{ravfogel2023conformal}, selective temperature control for multi-sample inference \cite{du2025optimizing,troshin2025selective}, and control-theoretic decoding in pre-logit space \cite{kanai2025aisp}.

\noindent
\textbf{Reasoning Compute Control.} Reasoning compute is another major inference-time factor. Chain-of-thought \cite{wei2022chain}, self-consistency \cite{wang2023selfconsistency}, and tree-structured deliberation \cite{yao2024tot} established that allocating extra reasoning steps or sampled reasoning paths can substantially improve model performance. Recent models and scaling analyses further frame test-time compute as a primary lever for performance, especially on mathematics, code, and other reasoning-intensive domains \cite{openai2024o1,snell2024scaling}. At the same time, a growing body of work from both academia and industry shows that longer reasoning is not uniformly better. Answers often converge before generation ends, and excessive deliberation can introduce redundancy or overthinking \cite{liu2025answerconvergence,wang2026rom,openai2025gpt5developers,li2025thinkbench,fan2026chainsseeanswersdont}. In response, existing methods study adaptive budget allocation \cite{lin2025planbudget,shi2026spader}, early stopping based on convergence or confidence signals \cite{yang2025deer}, and extra self-correction/refinement passes when additional reasoning appears useful \cite{madaan2023selfrefine,weng2023selfverification}.

\noindent
\textbf{Toward Within-Trajectory Joint Control.} While the above methods adapt a single inference-time dimension, recent work has begun to jointly adapt both. AdaReasoner \cite{wang2025adareasoner}, for example, selects question-tailored reasoning instruction format, temperature, and reasoning steps, while task-level inference optimization methods such as EcoTune \cite{xu2025ecotune} jointly tune hyperparameters like temperature and maximum output length. However, these methods still operate mainly at the request level. The controller chooses a configuration before generation, and the full rollout then follows that choice, so the trajectory itself remains largely uncontrolled. Recent analyses suggest that reasoning trajectories exhibit identifiable latent states and transition patterns rather than uniform behavior throughout a rollout \cite{sun2026trajectories,sun2026towards}, that temporal confidence, entropy evolution, and structural ordering within a chain carry richer diagnostic information than static summaries \cite{zhu2026edis,he2026order}, and that reasoning models often signal answer readiness well before they actually stop \cite{sharma2025thinkjustenough}. Other studies show that answer convergence and overthinking typically emerge only after partial progress has already been made, pointing to the importance of stage-sensitive intervention rather than one-shot configuration selection \cite{deng2023unifiedcalibration,chen2025drsaf}. Therefore, within-trajectory control is not merely a finer implementation, but a different granularity of inference-time decision. These observations motivate inference-time controllers that operate within a single reasoning trajectory, dynamically coordinating sampling stochasticity and reasoning budget.

\section{Within-Trajectory Joint Control}
\label{sec:view}

\paragraph{Definition.}
We adopt the standard view of LLM reasoning as an autoregressive generation process that produces a token sequence $\mathbf{x} = (x_1, x_2, \ldots, x_T)$, where each token is sampled via

\begin{equation}
    x_t \sim p_\theta\!\left(\cdot \mid x_{<t};\, \mathbf{c}_{s,t},\, \mathbf{c}_{c,t}\right),
    \label{eq:ar}
\end{equation}

with $\mathbf{c}_{s,t}$ and $\mathbf{c}_{c,t}$ denoting the control state at step $t$ along two dimensions: decoding stochasticity ($\mathbf{c}_{s,t}$), controlling \emph{how} the next token is drawn by parameterizing the shape of the sampling distribution, and reasoning compute ($\mathbf{c}_{c,t}$), controlling \emph{whether} generation proceeds by determining the continuation condition $\mathbb{1}[t < \tau]$ with $\tau$ the stopping time.

Let $\mathbf{c}_t \triangleq (\mathbf{c}_{s,t},\, \mathbf{c}_{c,t})$ collect both components.
A common inference-time practice is to operate at the \emph{request level}: $\mathbf{c}_t$ is fixed before generation begins and remains constant throughout the entire rollout.
We instead define \textbf{within-trajectory control} as a regime in which $\mathbf{c}_t$ is revised at a set of decision points $\mathcal{B} = \{t_1, t_2, \ldots\} \subset \{1, \ldots, T\}$ during generation, via
\begin{equation}
    \mathbf{c}_{t_k} = f\!\left(x_{<t_k};\, \mathbf{c}_{t_{k-1}}\right), \quad t_k \in \mathcal{B},
    \label{eq:wtc}
\end{equation}
while remaining unchanged between consecutive decision points.
The control is \emph{joint} in that the same controller $f$ updates both components of $\mathbf{c}_{t_k}$ simultaneously at each decision point, rather than selecting them independently or fixing one while adapting the other.

\paragraph{Control space completeness.}
We argue that our two-dimensional decomposition of $\mathbf{c}_t$ is not arbitrary. In the within-trajectory setting of LLM inference, it exhausts all control degrees of freedom. At every step $t$, the only degrees of freedom available to a controller are intervention on the shape of the sampling distribution ($\mathbf{c}_{s,t}$) and continuation condition ($\mathbf{c}_{c,t}$). Any concrete control parameter, including temperature, top-$p$, repetition penalties, budget limits, or early-stopping signals, instantiates one of these two components. No third structural class exists within a single evolving trajectory.

We note that multi-trajectory methods such as beam search or best-of-$n$ sampling do not constitute counterexamples \cite{freitag2017beamsearch,wang2025stbon}. They alter the inference topology by maintaining a set of concurrent trajectories, and therefore fall outside the \emph{single-trajectory} scope that defines our setting. A detailed argument for this completeness claim is provided in Appendix~\ref{app:completeness}.

\section{AutoCRAT}
\label{sec:autocrat}

\begin{figure*}[t]
\centering
\resizebox{0.95\textwidth}{!}{\includegraphics[trim=0 10 0 10,clip]{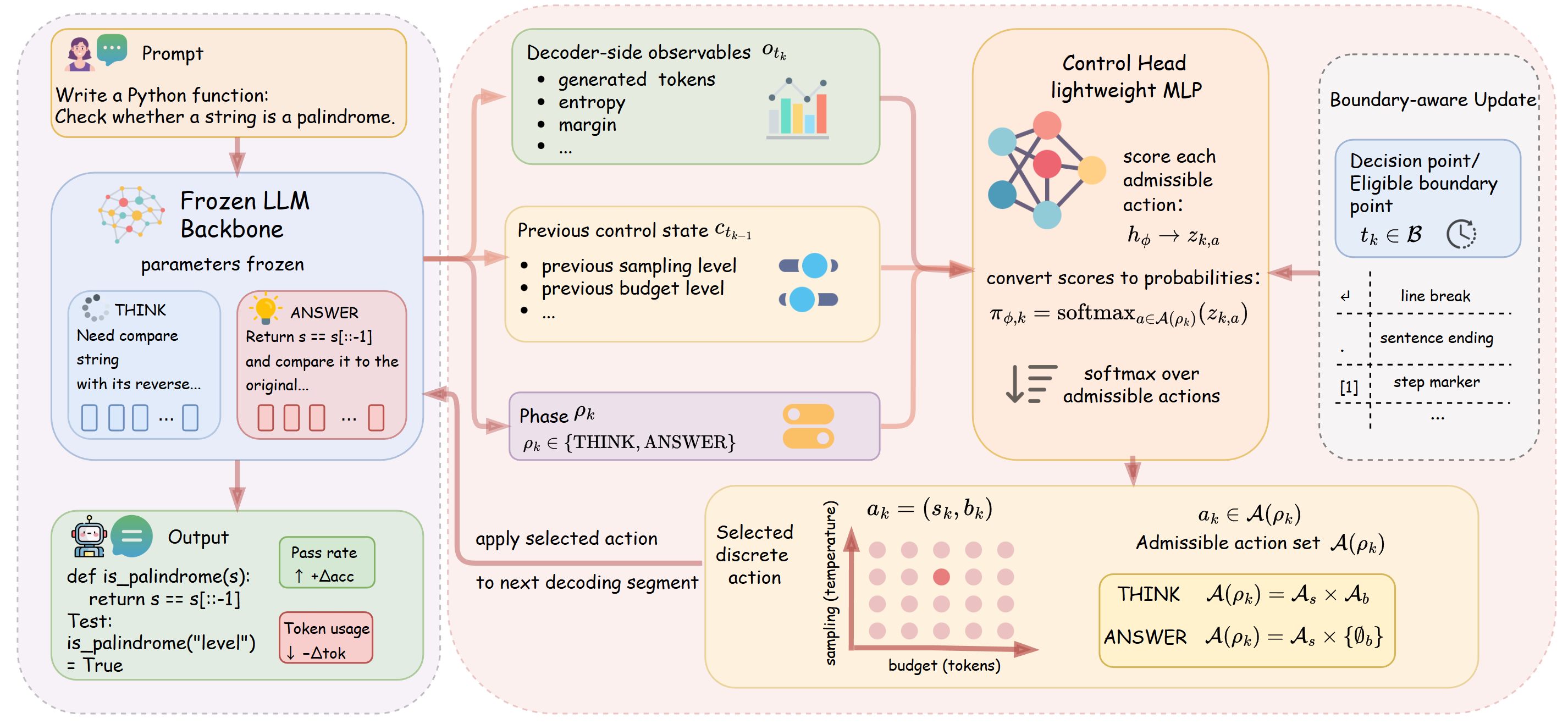}}
\caption{Overview of \textbf{\texttt{AutoCRAT}}.}
\label{fig:architecture}
\end{figure*}

Figure~\ref{fig:architecture} illustrates the architecture of our method. As \textbf{the first within-trajectory joint control framework}, \textbf{\texttt{AutoCRAT}} attaches to a frozen backbone without modifying its parameters or accessing its hidden states. A lightweight \emph{control head} maps decoder-side observables and previous control states to two discrete outputs, a \emph{sampling level} for sampling stochasticity and a \emph{budget level} for reasoning budget, which are updated jointly at semantic boundaries as generation unfolds. Section~\ref{ssec:head} details the control head design and action space. Section~\ref{ssec:boundary} describes the boundary-aware update mechanism. Section~\ref{ssec:training} presents the training procedure.

\subsection{Control Head and Action Space}
\label{ssec:head}

The control head $h_\phi$ is the decision-making component of \textbf{\texttt{AutoCRAT}}, implementing the control update in Eq.~\eqref{eq:wtc}. We denote the decoder-side observables at the decision point $t_k \in \mathcal{B}$ by $\mathbf{o}_{t_k}$, i.e., features derived from the decoding process without accessing backbone hidden states. The full specification is provided in Appendix~\ref{app:features}. At each $t_k$, $h_\phi$ maps $\mathbf{o}_{t_k}$ and the previous control state $\mathbf{c}_{t_{k-1}}$ to a categorical policy $\pi_{\phi,k}$ over control actions:
\begin{equation}
    h_\phi(\mathbf{o}_{t_k},\,\mathbf{c}_{t_{k-1}},\,\rho_k)
    = \pi_{\phi,k},
    \label{eq:control_head}
\end{equation}
where $\rho_k \in \{\textsc{think},\textsc{answer}\}$ denotes the current phase, and $\pi_{\phi,k}$ is a distribution over admissible control actions at $t_k$. $\mathbf{c}_{t_{k-1}}$ encodes the sampling and budget levels active at the preceding decision point, represented as concatenated one-hot vectors. An admissible action is written as $a_k=(s_k,b_k)$, where $s_k$ is the \emph{sampling level} instantiating $\mathbf{c}_{s,t_k}$, and $b_k$ is the \emph{budget level} instantiating $\mathbf{c}_{c,t_k}$. The action set is defined as:
\begin{equation}
    \mathcal{A}(\rho_k)=
    \begin{cases}
        \mathcal{A}_s \times \mathcal{A}_b, & \rho_k=\textsc{think},\\
        \mathcal{A}_s \times \{\varnothing_b\}, & \rho_k=\textsc{answer},
    \end{cases}
    \label{eq:phase_action_space}
\end{equation}
where $\mathcal{A}_s$ and $\mathcal{A}_b$ denote the admissible sampling and budget levels, and $\varnothing_b$ denotes a no-op budget action. Since the answer phase is not intended to contain explicit reasoning traces, reasoning-budget control is frozen while the sampling level remains active.

The control head is implemented as a lightweight MLP $g_\phi$ that takes the concatenation of $\mathbf{o}_{t_k}$ and $\mathbf{c}_{t_{k-1}}$ as input and produces a score for each admissible action $a = (s, b) \in \mathcal{A}(\rho_k)$, followed by a softmax normalization:
\begin{equation}
    \begin{aligned}
    \pi_{\phi,k}
    &=
    \operatorname{Softmax}_{a \in \mathcal{A}(\rho_k)}
    \!\left(z_{k,a}\right),\\
    z_{k,a}
    &=
    g_\phi\!\left([\mathbf{o}_{t_k};\,\mathbf{c}_{t_{k-1}}]\right)_a,
    \end{aligned}
    \label{eq:control_policy}
\end{equation}
where $z_{k,a}$ is the scalar score assigned to action $a$. The full pipeline is summarized in Algorithm~\ref{alg:control_workflow}.

\begin{algorithm}[H]
\caption{\textbf{\texttt{AutoCRAT}} Control Workflow}
\label{alg:control_workflow}
\small
\KwInput{Prompt $q$, frozen backbone $f_{\theta}$, control head $g_{\phi}$, boundary detector $\mathcal{B}$, initial state $(s_0,b_0)$}
\KwOutput{Final output $\hat{y}$ and boundary trace $\mathcal{M}$}
\AlgComment{Initialize controller state and trace}\;
$\mathbf{x}_{\le t_0}\!\leftarrow\!q,\ \mathbf{c}_{t_0}\!\leftarrow\!(s_0,b_0),\ \rho_0\!\leftarrow\!\textsc{think},\ \mathcal{M}\!\leftarrow\!\emptyset$\;
\While{not terminated}{
    \AlgComment{Decode local segment under active control}\;
    $\mathbf{x}_{(t_k,t_{k+1}]}\sim f_{\theta}(\cdot\mid \mathbf{x}_{\le t_k},\mathbf{c}_{t_k})$ \EqMark{eq:wtc}\;
    \If{$t_{k+1}\in\mathcal{B}$}{
        \AlgComment{Boundary-level control update}\;
        $\mathbf{o}_{t_{k+1}}\leftarrow \textsc{Obs}(\mathbf{x}_{\le t_{k+1}})$\;
        $\mathcal{A}_{k+1}\leftarrow\mathcal{A}(\rho_{k+1})$ \EqMark{eq:phase_action_space}\;
        $z_{k+1,a}\leftarrow g_{\phi}([\mathbf{o}_{t_{k+1}};\mathbf{c}_{t_k}])_a,\ \forall a\in\mathcal{A}_{k+1}$ \EqMark{eq:control_policy}\;
        $\pi_{\phi,k+1}(a)\leftarrow \dfrac{\exp(z_{k+1,a})}{\sum_{a'\in\mathcal{A}_{k+1}}\exp(z_{k+1,a'})},\ \forall a\in\mathcal{A}_{k+1}$ \EqMark{eq:control_policy}\;
        $a_{k+1}\sim \pi_{\phi,k+1},\ \mathbf{c}_{t_{k+1}}\leftarrow a_{k+1},\ \mathcal{M}\leftarrow\mathcal{M}\cup\{(t_{k+1},\mathbf{o}_{t_{k+1}},a_{k+1})\}$\;
    }
    \If{budget exhausted $\lor$ answer complete $\lor$ EOS}{
        \AlgComment{Boundary-triggered stop/phase transition}\;
        \textbf{break / phase-transition}\;
    }
}
\textbf{return} $\hat{y},\mathcal{M}$\;
\end{algorithm}

We use discrete levels rather than continuous values because the control head is driven by noisy decoder-side signals; discretization reduces high-frequency oscillation between consecutive boundaries and keeps each segment under a stable, interpretable configuration. All examples are initialized from a fixed balanced control state $(s_0,b_0)=(0.7,1024)$. This avoids task-specific routing and leaves subsequent adaptation to the evolving trajectory state. The concrete level assignments and decoding parameters are detailed in Section~\ref{ssec:setup}. The two dimensions are instantiated as follows:

\begin{itemize}[leftmargin=*,
    topsep=-0.5em,
    partopsep=0pt,
    parsep=0pt,
    itemsep=0pt]

\item[\ding{224}] \textbf{\textit{Sampling level}} 
 controls decoding stochasticity by selecting from a discrete set of sampling configurations. In the current instantiation of \textbf{\texttt{AutoCRAT}}, it is implemented via temperature only, as simultaneously adjusting multiple sampling parameters (e.g., \texttt{top\_p}, \texttt{top\_k}) introduces inter-parameter interactions that render the resulting sampling behavior difficult to predict or interpret, a practice explicitly discouraged by OpenAI and Anthropic \cite{openai2026responsesapi,anthropic2025apireleasenotes}. Notably, the sampling dimension is in principle compatible with other stochasticity controls. The current restriction ensures clean attribution of empirical gains to within-trajectory joint control, with each sampling level corresponding to a single, interpretable degree of stochasticity.

\item[\ding{224}] \textbf{\textit{Budget level}}
 imposes upper bounds on the total number of reasoning tokens, following prior work that frames token allocation as the primary lever for compute control~\cite{han2025tokenbudget,muennighoff2025s1}. Specifically, the bound is absolute, and when the active budget level is updated to a value whose bound falls below the tokens already consumed, the remaining allowance is immediately treated as exhausted, without revising previously generated tokens. Similar to the sampling level, the budget level is in principle compatible with other reasoning compute controls such as step-level pruning.

\end{itemize}

\subsection{Boundary-Aware Update}
\label{ssec:boundary}

To balance responsiveness and stability, \textbf{\texttt{AutoCRAT}} updates control decisions only at eligible semantic boundary points $t_k \in \mathcal{B}$, which are specified in Appendix~\ref{app:boundary_details}. Between two consecutive decision points, the selected action $(s_k,b_k)$ is held fixed for all generated tokens. At $t_k$, $h_\phi$ observes the updated $\mathbf{o}_{t_k}$ and $\mathbf{c}_{t_{k-1}}$, and may select a new action according to Eq.~\eqref{eq:control_head}. This boundary-level granularity avoids high-frequency token-level switching while aligning updates with step-like transitions in reasoning traces~\cite{liu2025answerconvergence}.

Boundary-aware updates are especially important for the budget control. Since budget reductions directly change the upper bounds and shorten the remaining reasoning phase, applying them in the middle of an unfinished fragment can create abrupt reasoning-answer transitions and unstable outputs~\cite{wu2025interruptible,zou2026userschangemindevaluating}. \textbf{\texttt{AutoCRAT}} instead applies a new budget level only after an eligible semantic boundary. If the lower budget makes the remaining reasoning allowance exhausted, reasoning is cut from that boundary onward. This keeps budget reductions interpretable by turning them into boundary-level reasoning termination. The controller may stop further reasoning, but only after the current local segment has been completed, avoiding arbitrary mid-fragment interruption. Backbones may still overflow a short residual segment while closing that fragment, which is why trajectories in Appendix~\ref{sec:appendix_trajectories} can show brief output under a zero remaining budget.

\subsection{Offline Training from Static Traces}
\label{ssec:training}
We train the control head $h_\phi$ fully offline from static traces, decoupling controller learning from online exploration noise and keeping supervision stable under a shared data distribution. Supervision is constructed in two stages. First, we collect trajectories from a fixed discrete action grid and convert raw runs into canonical traces with standardized fields, including task outcome, token cost, and ordered boundary events. Each boundary event is aligned to a decision point $t_k$ and represented as a boundary-level sample containing the local observable state and active control action. Second, we assign each trajectory a quality score that balances task correctness and normalized token cost:
\begin{equation}
    r(\tau)=\operatorname{Acc}(\tau)-\lambda \,\widetilde{C}(\tau),
    \label{eq:training_reward}
\end{equation}
where $\tau$ denotes a trajectory, $\operatorname{Acc}(\tau)$ is task correctness, and $\widetilde{C}(\tau)$ is problem-level normalized token cost, i.e., z-score within static trajectories of the same problem. $\lambda$ is a fixed scalar coefficient that governs the accuracy--efficiency tradeoff. A larger $\lambda$ shifts the controller toward efficiency-oriented behavior, making the framework adaptable to different deployment requirements. These scores are then used to construct trajectory-level preference pairs, and boundary-level supervision samples.

Given boundary state samples, the control head is trained with two complementary signals: behavior cloning for distribution matching over actions, and preference learning for pairwise ranking between better and worse actions in comparable local contexts. The training objective is
\begin{equation}
    \mathcal{L}_{\text{train}}
    =
    \mathcal{L}_{\text{BC}}
    +
    \beta \mathcal{L}_{\text{pref}},
    \label{eq:training_loss}
\end{equation}
with
\begin{equation}
    \mathcal{L}_{\text{BC}}
    =
    -\frac{1}{|\mathcal{K}|}
    \sum_{k\in\mathcal{K}}
    \sum_{a\in\mathcal{A}(\rho_k)}
    q_k(a)\log \pi_{\phi,k}(a),
    \label{eq:bc_loss}
\end{equation}
\begin{equation}
    \mathcal{L}_{\text{pref}}
    =
    \frac{1}{|\mathcal{P}|}
    \sum_{p\in\mathcal{P}}
    w_p \log\!\big(1+\exp(-\Delta z_p)\big),
    \label{eq:pref_loss}
\end{equation}
where $\mathcal{K}$ is the set of boundary samples and $\mathcal{P}$ is the set of preferred action pairs. Both supervision terms are derived from trajectory quality in Eq.~\eqref{eq:training_reward}: $q_k(a)$ is a reward-weighted soft target over actions under the same boundary context, while $w_p$ is a reward-gap weight for matched preference pairs. For $p=(k,a^+,a^-)$, $\Delta z_p=z_{k,a^+}-z_{k,a^-}$. Full construction is provided in Appendix~\ref{app:train_algorithm}.

For scale and transfer, we use a unified small-backbone pipeline for trace collection and control head training, then deploy the same controller on larger backbones without retraining. Under this offline formulation, the same trace pool can be reused to isolate transferable control dynamics from backbone-specific optimization.

\section{Experiments}
\label{sec:experiments}

\subsection{Experimental Setup}
\label{ssec:setup}

\begin{table*}[t]
\begin{center}
\resizebox{0.99\textwidth}{!}{%
\scriptsize
\def\arraystretch{1.05}
\setlength{\tabcolsep}{3.0pt}
\begin{tabular}{ll*{7}{cc}}
\toprule
\multirow{2}{*}{\scriptsize\bf Model} & \multirow{2}{*}{\scriptsize\bf Setting} & \multicolumn{2}{c}{\scriptsize\bf GSM8K} & \multicolumn{2}{c}{\scriptsize\bf MATH-500} & \multicolumn{2}{c}{\scriptsize\bf ARC-C} & \multicolumn{2}{c}{\scriptsize\bf GPQA-D} & \multicolumn{2}{c}{\scriptsize\bf HumanEval} & \multicolumn{2}{c}{\scriptsize\bf MBPP} & \multicolumn{2}{c}{\scriptsize\bf Avg.} \\
\cmidrule(r){3-4} \cmidrule(r){5-6} \cmidrule(r){7-8} \cmidrule(r){9-10} \cmidrule(r){11-12} \cmidrule(r){13-14} \cmidrule(r){15-16}
& & \scriptsize\bf Acc. & \scriptsize\bf Tok. & \scriptsize\bf Acc. & \scriptsize\bf Tok. & \scriptsize\bf Acc. & \scriptsize\bf Tok. & \scriptsize\bf Acc. & \scriptsize\bf Tok. & \scriptsize\bf Acc. & \scriptsize\bf Tok. & \scriptsize\bf Acc. & \scriptsize\bf Tok. & \scriptsize\bf Acc. & \scriptsize\bf Tok. \\
\midrule
\multirow{3}{*}{\texttt{Qwen3-4B}} & \textsc{Base} & 88.6 & 251 & 83.5 & 348 & 61.8 & 120 & 42.4 & 187 & 59.5 & 66 & 44.6 & 131 & 63.4 & 184 \\
& \textsc{Recommend} & 92.0 & 2189 & 94.5 & 2765 & 68.4 & 1042 & 52.5 & 1276 & 67.2 & 2665 & 52.5 & 1708 & 71.2 & 1941 \\
& \textbf{\texttt{AutoCRAT}} & \textbf{93.1} & \textbf{729} & \textbf{95.8} & \textbf{1333} & \textbf{70.2} & \textbf{382} & \textbf{53.8} & \textbf{1220} & \textbf{71.0} & \textbf{1138} & \textbf{55.4} & \textbf{1091} & \textbf{73.2} & \textbf{982} \\
\midrule
\multirow{3}{*}{\texttt{Qwen3-8B}} & \textsc{Base} & 91.0 & 258 & 87.5 & 366 & 68.2 & 151 & 47.5 & 244 & 65.6 & 67 & 47.8 & 235 & 67.9 & 220 \\
& \textsc{Recommend} & 95.8 & 2195 & 95.8 & 2692 & 76.4 & 1096 & 58.9 & 1492 & 71.8 & 2503 & 56.3 & 1715 & 75.8 & 1949 \\
& \textbf{\texttt{AutoCRAT}} & \textbf{96.8} & \textbf{562} & \textbf{97.0} & \textbf{1088} & \textbf{77.8} & \textbf{296} & \textbf{60.1} & \textbf{1375} & \textbf{75.6} & \textbf{864} & \textbf{59.5} & \textbf{1096} & \textbf{77.8} & \textbf{880} \\
\midrule
\multirow{3}{*}{\texttt{DSR1-8B}} & \textsc{Base} & 68.4 & 546 & 76.0 & 972 & 63.2 & 573 & 42.4 & 771 & 48.1 & 995 & 40.5 & 781 & 56.4 & 773 \\
& \textsc{Recommend} & 78.3 & 878 & 88.3 & 1484 & 66.2 & 643 & 48.1 & 1131 & 51.1 & 2978 & 46.0 & 1546 & 63.0 & 1443 \\
& \textbf{\texttt{AutoCRAT}} & \textbf{79.8} & \textbf{658} & \textbf{90.0} & \textbf{1243} & 65.9 & \textbf{604} & \textbf{49.4} & \textbf{1109} & \textbf{55.0} & \textbf{2195} & \textbf{50.7} & \textbf{1437} & \textbf{65.1} & \textbf{1208} \\
\midrule
\multirow{3}{*}{\texttt{Qwen3-30B}} & \textsc{Base} & 93.4 & 310 & 91.0 & 462 & 75.6 & 210 & 53.2 & 305 & 69.5 & 92 & 54.8 & 280 & 72.9 & 277 \\
& \textsc{Recommend} & 97.4 & 2520 & 97.5 & 3150 & 82.0 & 1180 & 65.2 & 1660 & 76.3 & 2840 & 62.2 & 1880 & 80.1 & 2205 \\
& \textbf{\texttt{AutoCRAT}} & \textbf{98.2} & \textbf{710} & \textbf{98.0} & \textbf{1260} & \textbf{83.5} & \textbf{340} & \textbf{66.5} & \textbf{1450} & \textbf{79.4} & \textbf{1010} & \textbf{64.8} & \textbf{1220} & \textbf{81.7} & \textbf{998} \\
\bottomrule
\end{tabular}
}
\end{center}
\vspace{-1em}
\caption{Main results across backbones and benchmarks. Acc. denotes accuracy for non-code tasks and pass@1 for code tasks, reported as percentages. Tok. denotes average completion tokens.}
\label{tab:main}
\end{table*}

\vspace{0.2em}
\noindent
\textbf{Benchmarks \& Metrics.}
We evaluate \textbf{\texttt{AutoCRAT}} on \textbf{6} benchmarks covering \textbf{3} reasoning-intensive domains: \textbf{(1) math reasoning}, GSM8K \cite{gsm8k} and MATH-500 \cite{lightman2023letsverify}; \textbf{(2) challenging QA}, ARC-C \cite{clark2018arc} and GPQA-D \cite{rein2023gpqa}; and \textbf{(3) code generation}, HumanEval \cite{human-eval} and MBPP \cite{austin2021mbpp}. For metrics, we report accuracy for non-code tasks and execution-based pass@1 under the provided unit tests for code tasks. For inference cost, we report the average number of completion tokens. The dataset statistics and split details are in Appendix~\ref{app:dataset}.

\vspace{0.2em}
\noindent
\textbf{Backbones \& Training Details.}
We leverage \textbf{4} open-source, instruct-tuned, and text-only models that natively produce explicit reasoning traces: Qwen3-4B (\texttt{Qwen3-4B}), Qwen3-8B (\texttt{Qwen3-8B}), Qwen3-30B-A3B (\texttt{Qwen3-30B})~\cite{qwen2025qwen3}, and DeepSeek-R1-Distill-Llama-8B (\texttt{DSR1-8B})~\cite{guo2025deepseek}. For Qwen3 series models, which expose a switchable thinking mode, we keep thinking mode enabled throughout all experiments. We collect static traces from \texttt{Qwen3-4B} on the pooled training splits of six benchmarks to train a single controller, then apply it to all backbones on held-out test splits without any benchmark- or backbone-specific retraining. This setup evaluates scale-up and cross-family transferability. The controller is trained with accuracy as the primary reward signal, rather than efficiency targets which lack a natural reference point. The pooled training set contains 981 problems and 19{,}620 trajectories, yielding 312{,}487 boundary-level supervision samples and 74{,}863 preference pairs. Training data details are provided in Appendix~\ref{sec:appendix_impl}.

\vspace{-0.5em}
\begin{table*}[t]
\begin{center}
\resizebox{0.99\textwidth}{!}{%
\scriptsize
\def\arraystretch{1.05}
\setlength{\tabcolsep}{2.0pt}
\begin{tabular}{l*{7}{cc}}
\toprule
\multirow{2}{*}{\scriptsize\bf Setting} & \multicolumn{2}{c}{\scriptsize\bf GSM8K} & \multicolumn{2}{c}{\scriptsize\bf MATH-500} & \multicolumn{2}{c}{\scriptsize\bf ARC-C} & \multicolumn{2}{c}{\scriptsize\bf GPQA-D} & \multicolumn{2}{c}{\scriptsize\bf HumanEval} & \multicolumn{2}{c}{\scriptsize\bf MBPP} & \multicolumn{2}{c}{\scriptsize\bf Avg.} \\
\cmidrule(r){2-3} \cmidrule(r){4-5} \cmidrule(r){6-7} \cmidrule(r){8-9} \cmidrule(r){10-11} \cmidrule(r){12-13} \cmidrule(r){14-15}
& \scriptsize\bf Acc. & \scriptsize\bf Tok. & \scriptsize\bf Acc. & \scriptsize\bf Tok. & \scriptsize\bf Acc. & \scriptsize\bf Tok. & \scriptsize\bf Acc. & \scriptsize\bf Tok. & \scriptsize\bf Acc. & \scriptsize\bf Tok. & \scriptsize\bf Acc. & \scriptsize\bf Tok. & \scriptsize\bf Acc. & \scriptsize\bf Tok. \\
\midrule
\textsc{Recommend} & 92.0 & 2189 & 94.5 & 2765 & 68.4 & 1042 & 52.5 & 1276 & 67.2 & 2665 & 52.5 & 1708 & 71.2\AccDrop{2.7\%} & 1941\TokOver{97.7} \\
\textsc{EDT} & 92.3 & 2105 & 95.0 & 2638 & 69.0 & 1018 & 53.2 & 1238 & 68.7 & 2520 & 54.3 & 1629 & 72.1\AccDrop{1.5\%} & 1858\TokOver{89.2} \\
\textsc{DEER} & 91.7 & 846 & 93.5 & 1450 & 67.1 & 336 & 51.3 & 1262 & 64.9 & 920 & 50.7 & 906 & 69.9\AccDrop{4.5\%} & 953\TokUnder{2.9} \\
\textbf{\texttt{AutoCRAT}} & \textbf{93.1} & \textbf{729} & \textbf{95.8} & \textbf{1333} & \textbf{70.2} & \textbf{382} & \textbf{53.8} & \textbf{1220} & \textbf{71.0} & \textbf{1138} & \textbf{55.4} & \textbf{1091} & \textbf{73.2} & \textbf{982} \\
\bottomrule
\end{tabular}
}
\end{center}
\vspace{-1em}
\caption{Comparison with existing adaptive baselines. {\color{GainGreen}$\downarrow$} marks the relative accuracy drop. {\color{GainGreen}$\uparrow$} marks baseline token overhead over \textbf{\texttt{AutoCRAT}}, and {\color{GainRed}$\downarrow$} marks baseline token reduction below \textbf{\texttt{AutoCRAT}}.}
\label{tab:adaptive}
\end{table*}

\vspace{0.5em}
\noindent
\textbf{Baselines \& Implementation Details.}
\textbf{\texttt{AutoCRAT}} operates over 4 temperature levels for sampling $\{0, 0.3, 0.7, 1.0\}$ and 5 reasoning token levels for budget $\{0, 64, 256, 1024, 4096\}$. Top-$p$ is fixed at $0.95$ as recommended by the backbone providers, while other decoding parameters use the defaults. For all comparisons, we keep the prompting and single-sample evaluation protocol identical. We compare against \textbf{5} groups of baselines. \ding{182} \textbf{Static baselines} include \textsc{Base} (temperature $0$) and \textsc{Recommend} (temperature $0.6$, following provider recommendations), both without budget constraint. \ding{183} \textbf{Sampling control}: \textsc{EDT}~\citep{zhang2024edt} adjusts temperature via an entropy-based rule. \ding{184} \textbf{Budget control}: \textsc{DEER}~\citep{yang2025deer} controls reasoning compute through dynamic early exit. \ding{185} \textbf{Request-level joint control}: \textsc{AdaReasoner}~\citep{wang2025adareasoner} selects a question-tailored configuration of reasoning instruction format, temperature, and step limit before generation. \ding{186} \textbf{Test-time compute}: \textsc{Self-Consistency}~\citep{wang2023selfconsistency} aggregates five independent trajectories via majority voting. These baselines span the spectrum from static configuration through single-axis adaptive control to request-level joint tuning and multi-trajectory aggregation. \textsc{EDT} and \textsc{DEER} are compared on all six benchmarks (Table~\ref{tab:adaptive}); \textsc{AdaReasoner} and \textsc{Self-Consistency} are evaluated on the non-code subset (MATH-500, ARC-C, GPQA-D; Table~\ref{tab:stronger_baselines}), as \textsc{AdaReasoner}'s released implementation does not support execution-based code evaluation. Reproduction details are in Appendix~\ref{app:adaptive_reproduction}. We additionally include a static joint-control grid in Section~\ref{ssec:joint_analysis}.

\subsection{Experimental Results}
\label{ssec:main_results}

We structure the empirical analysis around \textbf{4} key \textbf{Obs}ervations, which systematically reveal the advantages of within-trajectory joint control and the design choices of \textbf{\texttt{AutoCRAT}}.  The main results cover all \textbf{4} backbones and \textbf{6} benchmarks. The adaptive methods comparison and subsequent analyses are all conducted on \texttt{Qwen3-4B} to isolate effects from cross-backbone transfer, with the adaptive methods comparison spanning all \textbf{6} benchmarks and the joint control and ablation analyses over three representative benchmarks. All reported numbers are evaluated on held-out test splits.

\textbf{Obs.\ding{182} \textbf{\texttt{AutoCRAT}} improves the accuracy--compute tradeoff over static baselines on varying backbones without retraining.} According to Table~\ref{tab:main}, \textbf{\texttt{AutoCRAT}} achieves a better accuracy--compute tradeoff than both static baselines across benchmarks, as any fixed configuration cannot be optimal across problems of varying difficulty and reasoning stages. Compared with \textsc{Recommend}, \textbf{\texttt{AutoCRAT}} improves average accuracy by $2.2$--$4.0\%$ while reducing token usage by $13.8\%$--$52.7\%$ across all four backbones. The controller trained on \texttt{Qwen3-4B} retains an overall advantage on all backbones. As the control head operates solely on decoder-side observables that are structurally comparable across architectures, the learned policy captures transferable reasoning dynamics rather than backbone-specific patterns. Table~\ref{tab:significance} further shows statistically consistent gains of \textbf{\texttt{AutoCRAT}} over \textsc{Recommend} across backbones. Multi-seed stability analysis in Appendix~\ref{app:seeds} confirms that these gains are consistent across three independent controller-training seeds.

\begin{table}[H]
    \centering
    \resizebox{0.99\columnwidth}{!}{%
    \scriptsize
    \def\arraystretch{1.05}
    \setlength{\tabcolsep}{3.5pt}
    \begin{tabular}{lccc}
    \toprule
    \textbf{Backbone} & \textbf{Wins} & \textbf{$\Delta$Acc.} & \textbf{Avg. Tok. Red.} \\
    \midrule
    \texttt{Qwen3-4B}  & 6/6 & +3.1\% [+1.9\%, +4.5\%] & 46.6\% \\
    \texttt{Qwen3-8B}  & 6/6 & +2.9\% [+1.6\%, +4.1\%] & 52.7\% \\
    \texttt{DSR1-8B}   & 5/6 & +4.0\% [+1.5\%, +5.6\%] & 13.8\% \\
    \texttt{Qwen3-30B} & 6/6 & +2.2\% [+1.1\%, +3.4\%] & 52.5\% \\
    \bottomrule
    \end{tabular}
    }
    
    \caption{Sample-level bootstrap summary against \textsc{Recommend}. $\Delta$Acc. reports the relative accuracy gain normalized by \textsc{Recommend} accuracy, with a 95\% confidence interval estimated by bootstrapping test examples within each benchmark and then averaging benchmark-level gains. Wins report the number of benchmarks where \textbf{\texttt{AutoCRAT}} improves accuracy. Avg. Tok. Red. reports the average relative token reduction.}
    \label{tab:significance}
    \end{table}

\textbf{Obs.\ding{183} \textbf{\texttt{AutoCRAT}} achieves a favorable accuracy--compute tradeoff against adaptive and request-level baselines.} As shown in Table~\ref{tab:adaptive}, compared to the single-axis controllers, \textbf{\texttt{AutoCRAT}} achieves higher average accuracy across all six benchmarks. \textbf{\texttt{AutoCRAT}} outperforms \textsc{EDT} by $1.5\%$ in relative accuracy while using $47.1\%$ fewer tokens. \textsc{DEER} trails \textbf{\texttt{AutoCRAT}} by $4.5\%$ in relative accuracy while using $2.9\%$ fewer tokens on average, suggesting that single-axis compute control can over-compress reasoning. Table~\ref{tab:stronger_baselines} further compares \textbf{\texttt{AutoCRAT}} with request-level and test-time compute methods on the non-code benchmarks. \textsc{AdaReasoner} matches \textbf{\texttt{AutoCRAT}} in average accuracy ($73.3$ vs.\ $73.3$) but consumes $73.1\%$ more tokens; \textsc{Self-Consistency} achieves comparable accuracy ($73.2$) at over $8{\times}$ the token cost. These results indicate that within-trajectory joint control attains a competitive accuracy level with substantially lower token consumption than both request-level joint tuning and multi-trajectory aggregation, the latter also facing known robustness challenges \cite{liu2026consensustraprescuingmultiagent}.

\begin{table}[H]
\centering
\resizebox{0.99\columnwidth}{!}{%
\scriptsize
\def\arraystretch{1.05}
\setlength{\tabcolsep}{3.0pt}
\begin{tabular}{l*{4}{cc}}
\toprule
\multirow{2}{*}{\scriptsize\bf Method} & \multicolumn{2}{c}{\scriptsize\bf MATH-500} & \multicolumn{2}{c}{\scriptsize\bf ARC-C} & \multicolumn{2}{c}{\scriptsize\bf GPQA-D} & \multicolumn{2}{c}{\scriptsize\bf Avg.} \\
\cmidrule(r){2-3} \cmidrule(r){4-5} \cmidrule(r){6-7} \cmidrule(r){8-9}
& \scriptsize\bf Acc. & \scriptsize\bf Tok. & \scriptsize\bf Acc. & \scriptsize\bf Tok. & \scriptsize\bf Acc. & \scriptsize\bf Tok. & \scriptsize\bf Acc. & \scriptsize\bf Tok. \\
\midrule
\textsc{AdaReasoner} & 96.0 & 2437 & 69.5 & 947 & \textbf{54.4} & 1694 & \textbf{73.3} & 1693 \\
\textsc{Self-Consistency} & \textbf{96.5} & 13187 & 69.8 & 5023 & 53.2 & 6184 & 73.2 & 8131 \\
\textbf{\texttt{AutoCRAT}} & 95.8 & \textbf{1333} & \textbf{70.2} & \textbf{382} & 53.8 & \textbf{1220} & \textbf{73.3} & \textbf{978} \\
\bottomrule
\end{tabular}
}
\caption{Comparison with request-level joint-control and test-time compute methods on \texttt{Qwen3-4B}. \textsc{Self-Consistency} aggregates five trajectories; its token cost sums all trajectories.}
\label{tab:stronger_baselines}
\end{table}

\subsection{Joint Control Analysis}
\label{ssec:joint_analysis}

To verify that the gains of \textbf{\texttt{AutoCRAT}} are indeed attributable to within-trajectory joint control, we compare against a static grid over the complete action space, which includes all possible fixed joint configurations. The oracle serves as an upper fixed-configuration reference.

\begin{table}[H]
\centering
\resizebox{0.99\columnwidth}{!}{%
\scriptsize
\def\arraystretch{1.05}
\setlength{\tabcolsep}{3.2pt}
\begin{tabular}{l cc cc cc}
\toprule
\multirow{2}{*}{}
& \multicolumn{2}{c}{\scriptsize\bf Oracle Config}
& \multicolumn{2}{c}{\scriptsize\bf Oracle}
& \multicolumn{2}{c}{\scriptsize\bf \textbf{\texttt{AutoCRAT}}} \\
\cmidrule(r){2-3} \cmidrule(r){4-5} \cmidrule(r){6-7}
& \scriptsize\bf Temp. & \scriptsize\bf Budget
& \scriptsize\bf Acc. & \scriptsize\bf Tok.
& \scriptsize\bf Acc. & \scriptsize\bf Tok. \\
\midrule
GSM8K     & 0.7 & 1024 & 92.5 & 1008 & 93.1\AccGain{0.6\%} & 729\TokGain{27.7} \\
ARC-C     & 0.7 & 1024 & 69.4 & 812 & 70.2\AccGain{1.2\%} & 382\TokGain{53.0} \\
HumanEval & 0.7 & 4096 & 68.7 & 2150 & 71.0\AccGain{3.3\%} & 1138\TokGain{47.1} \\
\midrule
Avg.      & --  & --   & 76.9 & 1323 & 78.1\AccGain{1.6\%} & 750\TokGain{43.4} \\
\bottomrule
\end{tabular}
}
\caption{Within-trajectory joint control versus oracle fixed joint control. Oracle selects the highest-accuracy fixed temperature--budget pair from the same action grid for each benchmark. {\color{GainRed}$\uparrow$} marks the relative accuracy gain of \textbf{\texttt{AutoCRAT}} over Oracle, and {\color{GainRed}$\downarrow$} marks the token reduction relative to Oracle.}
\label{tab:joint_control}
\end{table}

\begin{figure}[H]
\centering
\resizebox{0.99\columnwidth}{!}{\includegraphics{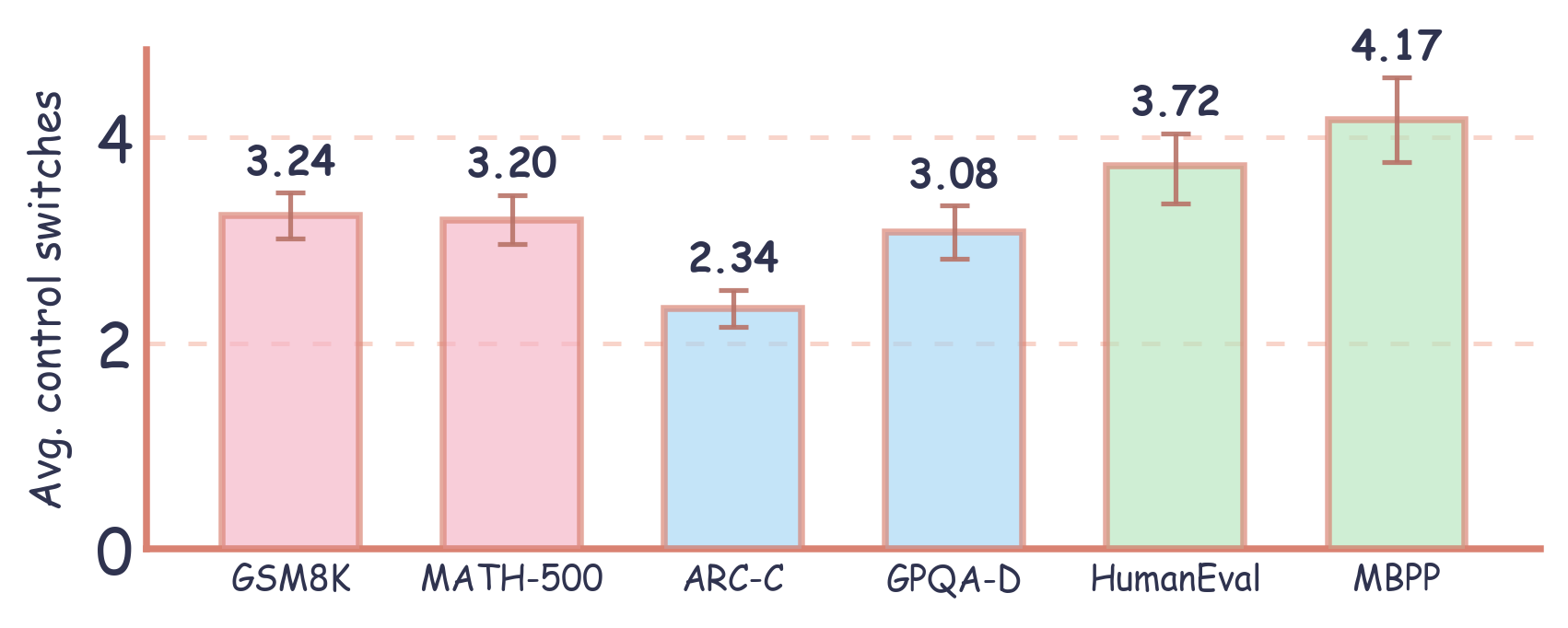}}
\caption{Average control-switching counts across six benchmarks.}
\label{fig:switching}
\end{figure}

\textbf{Obs.\ding{184} Within-trajectory joint control improves over oracle fixed joint control.} As shown in Table~\ref{tab:joint_control}, even when the fixed baseline is allowed to choose the highest-accuracy temperature--budget pair separately for each benchmark, \textbf{\texttt{AutoCRAT}} still improves the average accuracy by $1.6\%$, and uses 43.4\% fewer tokens. The gains are not caused by a one-shot configuration choice. Figure~\ref{fig:switching} shows that \textbf{\texttt{AutoCRAT}} actively revises its control state within trajectories, with average switching counts ranging from $2.34$ on ARC-C to $4.17$ on MBPP. The benchmark-level ranges further show that the switching behavior is task dependent: ARC-C has the lowest and tightest interval ($2.16$--$2.52$), while code tasks require more frequent and more variable control updates, reaching $3.36$--$4.04$ on HumanEval and $3.76$--$4.58$ on MBPP. This pattern indicates that different domains affect both the absolute number of control updates and their variability. Appendix~\ref{sec:appendix_trajectories} provides an intuitive view through representative controlled trajectories, showing how \textbf{\texttt{AutoCRAT}} works at different reasoning stages.

\subsection{Ablation Studies}
\label{ssec:ablations}

We conduct ablation studies to analyze the impact of different components of \textbf{\texttt{AutoCRAT}}.

\begin{table}[H]
\centering
\resizebox{0.99\columnwidth}{!}{%
\scriptsize
\def\arraystretch{1.05}
\setlength{\tabcolsep}{3.5pt}
\begin{tabular}{l*{3}{cc}}
\toprule
\multirow{2}{*}{\small\bf Setting} & \multicolumn{2}{c}{\scriptsize\bf GSM8K} & \multicolumn{2}{c}{\scriptsize\bf ARC-C} & \multicolumn{2}{c}{\scriptsize\bf HumanEval} \\
\cmidrule(r){2-3} \cmidrule(r){4-5} \cmidrule(r){6-7}
& \scriptsize\bf Acc. & \scriptsize\bf Tok. & \scriptsize\bf Acc. & \scriptsize\bf Tok. & \scriptsize\bf Acc. & \scriptsize\bf Tok. \\
\midrule
\textsc{Base} & 88.6 & 251 & 61.8 & 120 & 59.5 & 66 \\
\textsc{Recommend} & 92.0 & 2189 & 68.4 & 1042 & 67.2 & 2665 \\
Sampling-only & 92.2 & 1714 & 68.8 & 824 & 68.7 & 2192 \\
Budget-only & 91.9 & 668 & 67.2 & 316 & 66.4 & 987 \\
Token-level & 92.6 & 1019 & 69.1 & 451 & 68.7 & 1484 \\
\textbf{\texttt{AutoCRAT}} & \textbf{93.1} & \textbf{729} & \textbf{70.2} & \textbf{382} & \textbf{71.0} & \textbf{1138} \\
\bottomrule
\end{tabular}
}
\caption{Ablation results on \texttt{Qwen3-4B} over 3 representative benchmarks. Sampling-only and Budget-only adapt a single control axis, and Token-level updates the controller at every token rather than at semantic boundaries.}
\label{tab:ablation}
\end{table}

\textbf{Obs.\ding{185} Both joint control and boundary-aware updates are 
crucial design choices.} As shown in Table~\ref{tab:ablation}, \textbf{\texttt{AutoCRAT}} outperforms both sampling-only and budget-only variants in accuracy while achieving a better token-accuracy tradeoff. Compared with the sampling-only variant, \textbf{\texttt{AutoCRAT}} improves accuracy by $1.0$--$3.3\%$ across three benchmarks while reducing token usage by an average of $52.5\%$, confirming that budget control is not redundant given adaptive sampling. Compared with the budget-only variant, \textbf{\texttt{AutoCRAT}} achieves $1.3$--$6.9\%$ accuracy gains at only a marginal token overhead, demonstrating that sampling control recovers task performance at a modest extra cost. These results confirm that the two dimensions interact and their joint optimum cannot be recovered by either alone. Boundary-aware updates also prove their importance. Token-level switching yields lower accuracy and higher token use across all benchmarks, as frequent updates on noisy local signals disrupt reasoning fragments mid-completion, whereas semantic boundaries provide a more stable observation window over complete local units. Together, the studies show the complementary value of joint control and boundary-aware updates.

\subsection{Generalization}
\label{ssec:generalization}

To test whether the learned control policy generalizes beyond its training distribution, we directly apply the existing \texttt{Qwen3-4B}-trained controller to two unseen benchmarks, SVAMP~\cite{patel2021svamp} (1{,}000 questions) and CommonsenseQA~\cite{talmor2019commonsenseqa} (1{,}221 questions), without additional trace collection or controller retraining. Results are reported in Table~\ref{tab:ood}.

\begin{table}[H]
\centering
\resizebox{0.99\columnwidth}{!}{%
\scriptsize
\def\arraystretch{1.05}
\setlength{\tabcolsep}{3.5pt}
\begin{tabular}{ll*{2}{cc}}
\toprule
\multirow{2}{*}{\scriptsize\bf Backbone} & \multirow{2}{*}{\scriptsize\bf Setting} & \multicolumn{2}{c}{\scriptsize\bf SVAMP} & \multicolumn{2}{c}{\scriptsize\bf CommonsenseQA} \\
\cmidrule(r){3-4} \cmidrule(r){5-6}
& & \scriptsize\bf Acc. & \scriptsize\bf Tok. & \scriptsize\bf Acc. & \scriptsize\bf Tok. \\
\midrule
\multirow{2}{*}{\texttt{Qwen3-4B}} & \textsc{Recommend} & 81.3 & 1579 & 74.8 & 1083 \\
& \textbf{\texttt{AutoCRAT}} & \textbf{82.9} & \textbf{967} & \textbf{75.4} & \textbf{684} \\
\midrule
\multirow{2}{*}{\texttt{Qwen3-8B}} & \textsc{Recommend} & 86.1 & 1687 & 80.7 & 1136 \\
& \textbf{\texttt{AutoCRAT}} & \textbf{87.2} & \textbf{1039} & \textbf{81.3} & \textbf{649} \\
\bottomrule
\end{tabular}
}
\caption{Out-of-distribution generalization on unseen benchmarks. The controller is trained only on the original six benchmarks and applied without retraining.}
\label{tab:ood}
\end{table}

Across all four settings, \textbf{\texttt{AutoCRAT}} improves both accuracy and token efficiency on benchmarks entirely absent from its training data. The accuracy gains range from $0.6$ to $1.6$ points while token reductions reach $36.8$--$42.9\%$, confirming that the controller captures transferable reasoning dynamics rather than benchmark-specific patterns.

\section{Conclusion}
\label{sec:conclusion}
In this paper, we shift the view of inference-time LLM reasoning control from isolated or per-request adaptation to within-trajectory joint control of decoding stochasticity and reasoning compute. Building on this concept, we propose \textbf{\texttt{AutoCRAT}}, which dynamically controls both dimensions in a single trajectory, and achieves a substantially better accuracy--compute tradeoff than baselines. We believe that \textbf{\texttt{AutoCRAT}} paves the way toward more comprehensive, principled and effective inference-time control for LLM reasoning.

\section*{Limitations}

The central claim of this work is the value of within-trajectory joint control of decoding stochasticity and reasoning compute. To ensure clean attribution, we deliberately restrict the action space to temperature and token-budget limits and train the controller offline from static traces rather than online reinforcement learning. Richer actions (e.g., top-p, rollback) and online training are left for future work. A consequence of offline training is distribution shift: the controller learns from static-configuration trajectories but is deployed under adaptive control where its own decisions shape the trajectory distribution, potentially limiting the performance ceiling of the current approach. Iterative trace collection under the learned policy could mitigate this gap. The controller's reliance on surface-level decoder signals also makes it susceptible to false convergence, where fluent but incorrect reasoning exhibits low uncertainty and triggers premature budget reduction (Appendix~\ref{app:case_premature}).

For design tradeoffs, \textbf{\texttt{AutoCRAT}} relies solely on decoder-side observables and control-state variables. While this reduces coupling to specific model architectures, it also limits the access to finer-grained internal reasoning signals \cite{li2026toward,huang2026rethinkingmemorymechanismsfoundation}. Furthermore, boundary detection for update relies on lightweight structural heuristics, which may be less robust in tasks with weak or irregular discourse structure. Finally, we do not perform per-backbone retraining or online reinforcement learning. While these strategies may further improve absolute performance, they introduce backbone-specific optimization and are outside our transferability-focused scope.

The controller also introduces a non-zero per-boundary inference overhead, as the control head is invoked once at every eligible boundary. In our setting this overhead remains modest, since the control head is a lightweight  MLP over low-dimensional decoder-side features and the number of boundary invocations per trajectory remains bounded by a small constant in our experiments, well below the number of generated tokens. Nevertheless, on latency-sensitive deployments or with very long trajectories the cumulative overhead becomes non-trivial, and tighter co-design between the controller and the decoding loop is a worthwhile direction.

Ultimately, our experiments are currently limited to text-based reasoning on English benchmarks with accuracy-focused evaluation \cite{sun2026accuracymeasuringbiasacknowledgment}. The behavior of \textbf{\texttt{AutoCRAT}} or similar frameworks in broader interactive environments, including multilingual generalization \cite{gao2026laobench,wang2025mucar}, multimodal reasoning \cite{kang2026hssbench,kang2026multimodal}, and agentic workflows \cite{luo2026centaurevalbenchmarkinghumanintheloopvalue,xu2026masdrift} remains to be established.

\section*{Ethics Statement}
This work uses publicly available benchmarks and open models within their applicable licenses, terms, and intended research settings. It does not involve human-subject experiments, private data collection, or sensitive personal data annotation. Therefore, the ethical considerations of this study are limited and mainly concern compliance with dataset and model usage policies, together with responsible deployment to reduce the risk of misuse \cite{zhang2025agent,luo2026agentauditor}. Our evaluation is currently limited to English text benchmarks, and broader generalization is not assessed in this work.

\section*{Acknowledgements}
This work is supported in part by the NYUAD Center for Interdisciplinary Data Science \& AI (CIDSAI), funded by Tamkeen under the NYUAD Research Institute Award CG016.

\bibliography{custom}

\clearpage
\appendix

\section{Detail of Control Space Completeness}
\label{app:completeness}

This section provides a formal argument for the control-space completeness claim in Section~\ref{sec:view}, which asserts that the two-dimensional decomposition $\mathbf{c}_t = (\mathbf{c}_{s,t}, \mathbf{c}_{c,t})$ exhausts all control degrees of freedom within a single LLM inference trajectory. Appendix \ref{app:completeness:scope} \& \ref{app:completeness:interventions} formalize the admissible intervention space. Appendix \ref{app:completeness:proposition} states a completeness proposition. Appendix \ref{app:completeness:counterexamples} explains why potential counterexamples do not violate the claim.

\subsection{Scope of the Argument}
\label{app:completeness:scope}

The completeness claim is relative to a precisely defined setting. We consider \textbf{single-trajectory autoregressive decoding} (subsuming standard LLM reasoning and generation) with a \textbf{frozen backbone}. Under this setting, the controller is constrained as follows:

\begin{itemize}[
    leftmargin=1.6em,
    labelsep=0.5em,
    topsep=-0.5em,
    partopsep=0pt,
    parsep=0pt,
    itemsep=0pt
]
    \item[\ding{182}] not modify model parameters;
    \item[\ding{183}] not maintain or branch into multiple parallel candidate trajectories;
    \item[\ding{184}] not revise or retract previously generated tokens (generation is strictly forward);
    \item[\ding{185}] not introduce information from outside the current trajectory, including external tools, auxiliary verifiers, or reward models.
\end{itemize}

\vspace{0.5em}
We emphasize that completeness here refers to \emph{completeness of control degrees of freedom within this setting}. Methods that operate outside these constraints are not counterexamples to our claim.

\subsection{Admissible Interventions}
\label{app:completeness:interventions}

In standard autoregressive generation, the system at step $t$, given the token history $x_{<t}$, faces exactly two structural questions:
\begin{itemize}[leftmargin=*, topsep=2pt, itemsep=1pt]
    \item \textbf{Q1 (Continuation):} Should generation proceed, or should it stop?
    \item \textbf{Q2 (Sampling):} If generation proceeds, from what distribution should the next token $x_t$ be drawn?
\end{itemize}

These two questions admit the following formal objects of control. The continuation decision is parameterized by a stopping condition $\mathbb{1}[t < \tau]$, where $\tau$ denotes a possibly random stopping time. A controller controls $\tau$ dynamically. The sampling decision is parameterized by the (possibly controller-modified) next-token distribution

\begin{equation}
    \tilde{p}(x_t \mid x_{<t},\, \mathbf{c}_{s,t}) = \mathcal{T}\!\left[p_\theta(\cdot \mid x_{<t});\, \mathbf{c}_{s,t}\right],
\end{equation}

where $\mathcal{T}[\,\cdot\,]$ denotes any transformation of the base model's next-token distribution, instantiated for example by temperature scaling, nucleus truncation, logit biasing, or constrained decoding masks.

These two objects align precisely with $\mathbf{c}_{c,t}$ (continuation control) and $\mathbf{c}_{s,t}$ (distribution shaping). Throughout the paper we use decoding stochasticity control as a convenient shorthand for $\mathbf{c}_{s,t}$. It refers to interventions that shape the conditional next-token distribution, including but not limited to changes in randomness level. Intuitively, $\mathbf{c}_{s,t}$ controls how the trajectory explores at each step, while $\mathbf{c}_{c,t}$ controls how far the trajectory extends. Together, these two dimensions completely characterize the degrees of freedom of a single trajectory.

\subsection{Proposition}
\label{app:completeness:proposition}

\begin{proposition}[Control Space Completeness]
Under the single-trajectory setting, any admissible inference-time intervention that alters the rollout distribution does so through one of the following two pathways, or their combination:
\begin{itemize}[leftmargin=*, topsep=2pt, itemsep=1pt]
    \item[(i)] \textbf{Distribution shaping} $(\mathbf{c}_{s,t})$: modification of the conditional next-step token-selection distribution $\tilde{p}(x_t \mid x_{<t}, \mathbf{c}_{s,t})$;
    \item[(ii)] \textbf{Continuation control} $(\mathbf{c}_{c,t})$: modification of the stopping condition $\mathbb{1}[t < \tau]$.
\end{itemize}
No third intervention class exists whose effect on the rollout distribution cannot be reduced to, or expressed as a combination of, (i) and (ii). Certain implementations may couple both dimensions, as discussed in Appendix~\ref{app:completeness:counterexamples}.
\end{proposition}

We establish the proposition in three steps.

\paragraph{Two-component decomposition.}

In a forward-only, single-trajectory autoregressive process, the per-step evolution at step $t$ is governed by a conditional transition law $K_t$. At this granularity, a transition only decides whether the trajectory continues and, if it continues, which next token is produced. Given the token history $x_{<t}$ and the current control state $\mathbf{c}_t$, this law can be decomposed into two components:
\begin{equation}
    \begin{aligned}
        K_t \equiv \Big(&
        \underbrace{P\!\left(\mathrm{continue} \mid x_{<t},\, \mathbf{c}_{t}\right)}_{\text{continuation component}}, \\
        &
        \underbrace{\tilde{p}\!\left(x_t \mid \mathrm{continue},\, x_{<t},\, \mathbf{c}_{t}\right)}_{\text{sampling component}}
        \Big).
    \end{aligned}
    \label{eq:kernel}
\end{equation}
We separate these two components by structural role: the first governs whether the trajectory continues, and the second governs which token is produced if it does. The same concrete control implementation may affect one or both components. This decomposition is not an empirical observation; it is a direct structural consequence of the autoregressive generation mechanism under the forward-only, no-branching, no-rollback constraints of our setting. Any controller that influences the trajectory must do so by modifying at least one component of $K_t$.

\paragraph{Pathway reduction.}

Any concrete control parameter operates through one of two pathways:
\begin{itemize}[leftmargin=*, topsep=2pt, itemsep=1pt]
    \item \textbf{Pathway (i)} --- affecting the sampling component: temperature, top-$p$, top-$k$, repetition penalties, logit biases, grammar masks, and entropy-regularized rescaling all instantiate the transformation $\mathcal{T}[\,\cdot\,]$ on the base logits. They differ in their parameterization but share the same role: shaping $\tilde{p}(x_t \mid x_{<t})$.
    \item \textbf{Pathway (ii)} --- affecting the continuation component: maximum token budget, dynamic budget allocation, convergence-based early stopping, answer-readiness gating, and confidence-triggered termination all instantiate modifications to $\tau$. They differ in their triggering logic but share the same role: controlling whether generation proceeds.
\end{itemize}
Mechanisms in this setting instantiate one or both of these pathways, corresponding to $\mathbf{c}_{s,t}$ and $\mathbf{c}_{c,t}$ respectively.

\paragraph{Exhaustiveness.}

Under this kernel definition, if an intervention leaves both components of $K_t$ unchanged for every reachable history, then every step uses the same transition law as the original rollout. The induced rollout distribution is therefore unchanged by recursive composition over $t$. Such an intervention has no observable effect on produced tokens or continuation decisions, so it is not a separate control degree of freedom in this setting.

\subsection{Counterexample Handling}
\label{app:completeness:counterexamples}

We now address four categories of inference-time methods that may appear to challenge the proposition. They are either outside the scope, or can be reduced to one or both of the two dimensions.

\paragraph{(a) Beam search, best-of-$n$, and tree-structured search.}
These methods maintain a set of concurrent candidate trajectories, operating over a multi-trajectory structure rather than a single evolving sequence. Their defining characteristic is therefore a change in the structure of inference itself, not a modification of control within a single trajectory.

\paragraph{(b) Rollback, backtracking, and self-refinement with token revision.}
Methods that allow a system to retract previously generated tokens and rewrite earlier segments of the trajectory violate the forward-only, single-trajectory assumption (Constraint 3 in Appendix~\ref{app:completeness:scope}). Once rollback is permitted, the trajectory is no longer a monotonically growing sequence and the single-trajectory framework no longer applies.

\paragraph{(c) Hidden-state steering and activation-level intervention.}
Methods that directly read or modify the backbone's intermediate hidden states introduce a deeper access channel into the generation process. However, in any transformer-based LLM architecture \cite{vaswani2017attention}, such interventions ultimately influence the trajectory through the next-token logit distribution. A modified hidden state must pass through the remaining layers and the unembedding matrix before it affects which token is sampled. The effect therefore still terminates at $\tilde{p}(x_t \mid x_{<t})$, placing hidden-state steering within Pathway (i). If the intervention also changes cached states and later steps, those effects appear as changes to the corresponding future $\tilde{p}(x_t \mid x_{<t})$ terms; the cross-step effect is a composition of step-wise transition changes rather than a new control dimension.

\paragraph{(d) EOS token biasing.}
Mechanistically, increasing the logit of the end-of-sequence token to encourage earlier termination is a modification of the next-token distribution (Pathway i). Functionally, it can serve as an indirect mechanism for continuation control (Pathway ii). It is therefore a coupled instance of distribution shaping and continuation control, which is already covered by the proposition rather than a counterexample.

\section{Decoder-side Observable List}
\label{app:features}

This section summarizes the used decoder-side observables in Table~\ref{tab:feature_categories}.

\begin{table*}[t]
\centering
\small
\resizebox{0.99\textwidth}{!}{%
\begin{tabular}{@{}p{0.24\textwidth}p{0.72\textwidth}@{}}
\toprule
\textbf{Category} & \textbf{Included Features} \\
\midrule
Uncertainty features & Entropy, margin, top-1 probability, top-2 probability, top-k probability mass, EOS probability, EOS rank. \\
\midrule
Progress and budget features & Generated-token progress ratio, remaining reasoning-budget ratio, segment progress ratio. \\
\midrule
Structural boundary features & Boundary-kind indicators, answer-zone flag, think-zone flag. \\
\midrule
Control-state features & Joint sampling-budget one-hot in think phase, sampling-level one-hot in answer phase. \\
\bottomrule
\end{tabular}%
}
\caption{Feature categories used by \textbf{\texttt{AutoCRAT}} at boundary-level decision points.}
\label{tab:feature_categories}
\end{table*}

\begin{table*}[t]
\centering
\small
\resizebox{0.99\textwidth}{!}{%
\begin{tabular}{@{}lll@{}}
\toprule
\textbf{Cue / event} & \textbf{Definition} & \textbf{Examples} \\
\midrule
\multicolumn{3}{@{}l}{\textit{Semantic boundary:}} \\[2pt]
\quad Line break & Line- or paragraph-level separation & Newline, blank line \\
\quad Sentence ending & Sentence-final punctuation & Period, question mark, exclamation mark \\
\quad Step marker & Explicit reasoning-step delimiter & Numbered steps, \texttt{Step 1:}, \texttt{Reasoning:} \\[4pt]
\midrule
\multicolumn{3}{@{}l}{\textit{Termination event:}} \\[2pt]
\quad Natural EOS & Backbone-generated end-of-sequence event & EOS token \\
\quad Answer completion & Task-format or final-answer closure & Final-answer marker, code block or JSON closure \\
\quad Budget exhausted & Exhaustion of the active reasoning budget & Remaining reasoning allowance reaches zero \\
\bottomrule
\end{tabular}%
}
\caption{Boundary cues and termination events used by \textbf{\texttt{AutoCRAT}}. All cues are detected from generated text and decoding metadata.}
\label{tab:boundary_cues}
\end{table*}

\section{Boundary Detection Details}
\label{app:boundary_details}

Table~\ref{tab:boundary_cues} summarizes the boundary cues and termination events used by \textbf{\texttt{AutoCRAT}}. Semantic boundaries trigger regular control updates, and termination events end either the reasoning phase or the entire generation. Notably, \textsc{budget\_exhausted} only terminates the reasoning phase and transitions to the answer phase, instead of stopping generation. To avoid overly dense updates, a cue triggers a decision point only after a minimum number of 6 tokens from the previous one.

\section{Training Algorithm}
\label{app:train_algorithm}

This section presents the offline training workflow. Algorithm~\ref{alg:training_workflow} summarizes the full pipeline, from static-trace scoring and supervision construction, to joint optimization of behavior-cloning and preference losses \cite{shen2026fine}. The controller is trained once using pooled static traces collected from the training splits of all 6 benchmarks. For all experiments, we use the same trained controller without per-benchmark retraining. We define the boundary context key $\kappa_k$ by the same problem id, phase, boundary ordinal, and boundary cue type. \textsc{Ctx} returns this key; \textsc{MatchPairs} forms preference pairs only within the same key; and \textsc{TracePair} returns the higher- and lower-reward trajectories in such a pair.

\begin{algorithm}[!t]
\caption{Offline Training Workflow}
\label{alg:training_workflow}
\small
\KwInput{Static trajectories $\mathcal{T}$, coefficients $\lambda,\alpha,\gamma,\beta$, learning rate $\eta$}
\KwOutput{Trained controller parameters $\phi$}
\AlgComment{Canonicalize traces and assign trajectory quality}\;
\ForEach{$\tau\in\mathcal{T}$}{
    $r(\tau)\leftarrow\operatorname{Acc}(\tau)-\lambda\widetilde{C}(\tau)$ \EqMark{eq:training_reward}\;
}
\AlgComment{Build boundary samples and BC soft targets from $r(\tau)$}\;
$\mathcal{K}\leftarrow\textsc{CollectBoundarySamples}(\mathcal{T})$\;
\ForEach{$k\in\mathcal{K}$}{
    \ForEach{$a\in\mathcal{A}(\rho_k)$}{
        \AlgComment{Match traces that share boundary context $\kappa_k$ and choose action $a$}\;
        $\mathcal{T}_{k,a}\leftarrow\{\tau\in\mathcal{T}\mid \textsc{Ctx}(\tau,k)=\kappa_k,\ \textsc{Act}(\tau,k)=a\}$\;
        $q_{k,a}\leftarrow \sum_{\tau\in\mathcal{T}_{k,a}}\exp(\alpha r(\tau))$\;
    }
    $q_{k,a}\leftarrow q_{k,a}/\sum_{a'\in\mathcal{A}(\rho_k)}q_{k,a'},\ \forall a\in\mathcal{A}(\rho_k)$\;
}
\AlgComment{Build preference pairs and pair weights from reward gaps}\;
$\mathcal{P}\leftarrow\textsc{MatchPairs}(\mathcal{T},\mathcal{K})$\;
\ForEach{$p=(k,a^+,a^-)\in\mathcal{P}$}{
    $(\tau^+,\tau^-)\leftarrow\textsc{TracePair}(p),\ r(\tau^+)>r(\tau^-)$\;
    $w_p\leftarrow \sigma\!\big(\gamma(r(\tau^+)-r(\tau^-))\big)$\;
}
initialize $\phi$\;
\AlgComment{Joint optimization of BC and preference objectives}\;
\For{$e\leftarrow 1$ \KwTo $E$}{
    \ForEach{mini-batch $\mathcal{D}\subset\mathcal{K}$}{
        $z_{k,a}\leftarrow g_{\phi}([\mathbf{o}_{t_k};\mathbf{c}_{t_{k-1}}])_a,\ \forall a\in\mathcal{A}_k,\forall k\in\mathcal{D}$ \EqMark{eq:control_policy}\;
        $\mathcal{A}_k\leftarrow\mathcal{A}(\rho_k),\ \pi_{k,a}\leftarrow\operatorname{Softmax}_{a\in\mathcal{A}_k}(z_{k,a}),\ \forall k\in\mathcal{D}$ \EqMark{eq:control_policy}\;
        $\mathcal{L}_{\text{BC}}(\mathcal{D})\leftarrow-\frac{1}{|\mathcal{D}|}\sum_{k\in\mathcal{D}}\sum_{a\in\mathcal{A}_k}q_{k,a}\log\pi_{k,a}$ \EqMark{eq:bc_loss}\;
    }
    \ForEach{mini-batch $\mathcal{U}\subset\mathcal{P}$}{
        $\Delta z_p\leftarrow z_{k,a^+}-z_{k,a^-},\ \forall p=(k,a^+,a^-)\in\mathcal{U}$\;
        $\mathcal{L}_{\text{pref}}(\mathcal{U})\leftarrow\frac{1}{|\mathcal{U}|}\sum_{p\in\mathcal{U}}w_p\log(1+\exp(-\Delta z_p))$ \EqMark{eq:pref_loss}\;
    }
    \AlgComment{Use mini-batch averaged $\mathcal{L}_{\text{BC}}$ and $\mathcal{L}_{\text{pref}}$ in this epoch step}\;
    $\mathcal{L}_{\text{train}}\leftarrow\mathcal{L}_{\text{BC}}+\beta\mathcal{L}_{\text{pref}}$ \EqMark{eq:training_loss}\;
    $\phi\leftarrow\phi-\eta\nabla_{\phi}\mathcal{L}_{\text{train}}$\;
}
\textbf{return} $\phi$\;
\end{algorithm}

\section{Dataset Statistics}
\label{app:dataset}

Building upon established methodologies in workflow automation~\cite{saad2024archon,hu2024adas}, we divide each dataset into training and test sets using a \textsc{train:test} ratio of 1:4. The dataset statistics are included in Table~\ref{tab:dataset}. For ARC-Challenge, we pool the complete official train and test splits (2{,}291 questions in total) before resplitting, excluding only the 299-question development split. For evaluation, we adopt the official pipelines from the benchmarks. For non-code tasks, we report exact-match accuracy after task-specific answer extraction and normalization. For the mathematical tasks with symbolic answers in the non-code tasks, we additionally use the math grader when normalization alone is insufficient. For code tasks, we extract the final submitted program and report execution-based pass@1 under the provided unit tests. Other evaluation domains \cite{yu2025core3d,fu2025learning} are not covered.

\begin{table}[H]
\vspace{-0.5em}
\centering
\resizebox{0.99\columnwidth}{!}{
\begin{tabular}{l|cccc}
\toprule
Domain & Dataset & \#Train & \#Test & Metric  \\
\midrule
\multirow{2}{*}{Code Generation} & HumanEval & 33 & 131 & pass@1 \\
& MBPP & 86 & 341 & pass@1 \\
\midrule
\multirow{2}{*}{Math Reasoning} & GSM8K & 264 & 1055 & Acc \\
& MATH-500 & 100 & 400 & Acc \\
\midrule
\multirow{2}{*}{Challenging QA} & ARC-Challenge & 458 & 1833 & Acc \\
& GPQA-Diamond & 40 & 158 & Acc \\
\bottomrule
\end{tabular}}
\caption{Dataset Statistics.}
\label{tab:dataset}
\end{table}

\section{Adaptive Baseline Reproduction Details}
\label{app:adaptive_reproduction}

This section describes how we reproduce the two adaptive baselines used in Table~\ref{tab:adaptive}, \textsc{EDT}~\citep{zhang2024edt} and \textsc{DEER}~\citep{yang2025deer}. For tasks covered by the released implementations, we follow the original hyperparameter settings and do not perform per-benchmark calibration. For \textsc{DEER}, the public repository did not release the code-generation implementation at the time of our experiments; therefore, on HumanEval and MBPP we use a paper-faithful reimplementation of the same early-exit rule. We keep the original stopping threshold unchanged and do not tune it on code benchmarks. The stop parser is adapted only to recognize task-specific answer boundaries for code completion, while the early-exit decision rule itself is unchanged.

For both baselines, final answers, including code completions, are evaluated with the same task-specific extraction and scoring scripts used for \textbf{\texttt{AutoCRAT}}. Completion tokens are counted with the same tokenizer from the corresponding backbone. This keeps the comparison focused on whether a single-axis controller can match a within-trajectory joint controller under a shared evaluation pipeline.

For \textsc{AdaReasoner}~\citep{wang2025adareasoner}, we train one shared controller on 99 pooled questions (33 randomly sampled from each of MATH-500, ARC-C, and GPQA-D), with four training trajectories per question (396 in total). At test time, the controller selects a question-tailored configuration of reasoning instructions, temperature, and step limit, and one trajectory is generated per question. We preserve AdaReasoner's request-level action structure while adapting the task prompt and evaluation protocol to the three target benchmarks. Since AdaReasoner's official implementation does not support execution-based code evaluation, we evaluate all methods in the same non-code setting. For \textsc{Self-Consistency}~\citep{wang2023selfconsistency}, we generate five trajectories per question at temperature $0.6$ and aggregate via majority voting. Token cost sums all five trajectories.

\section{Training Details}
\label{sec:appendix_impl}

Table~\ref{tab:training_details} lists the controller training settings. The pooled training data statistics are summarized in Table~\ref{tab:training_scale}.

\begin{table}[H]
\centering
\small
\resizebox{0.99\columnwidth}{!}{
\begin{tabular}{l|r}
\toprule
\textbf{Stage} & \textbf{Count} \\
\midrule
Pooled training problems & 981 \\
Static trajectories ($4 \times 5$ grid) & 19{,}620 \\
Boundary-level supervision samples & 312{,}487 \\
Preference pairs & 74{,}863 \\
\bottomrule
\end{tabular}}
\caption{Training data scale.}
\label{tab:training_scale}
\end{table}

\begin{table}[H]
\centering
\small
\resizebox{0.99\columnwidth}{!}{
\begin{tabular}{l|l}
\toprule
\textbf{Setting} & \textbf{Value} \\
\midrule
Backbone & \texttt{Qwen3-4B} \\
Action grid & $4 \times 5$ configurations \\
Sampling levels & 0, 0.3, 0.7, 1.0 \\
Budget levels & 0, 64, 256, 1024, 4096 \\
Initial control state & $(0.7,1024)$ \\
Top-$p$ & $0.95$ \\
Minimum boundary interval & $6$ tokens \\
Overall budget & $16384$ tokens \\
Reward cost coefficient $\lambda$ & $0.2$ \\
Target softmax temperature & $0.25$ \\
MLP hidden dimension & $256$ \\
Dropout & $0.1$ \\
Optimizer & AdamW \\
Learning rate & $3\times10^{-4}$ \\
Weight decay & $10^{-4}$ \\
Pairwise loss weight & $0.5$ \\
Controller training epochs & $140$ \\
\bottomrule
\end{tabular}}
\caption{Fixed trace-collection and controller-training settings.}
\label{tab:training_details}
\end{table}

\section{Case Study of Controlled Trajectory}
\label{sec:appendix_trajectories}

This section provides two qualitative case studies that make the within-trajectory control concrete. Both examples use the \texttt{Qwen3-8B} backbone, and are selected from the original benchmarks. For readability, we report condensed boundary-level trajectories rather than raw token streams. The first example is a direct mathematical conversion from MATH-500, representing a case where the reasoning path is short and converges quickly. The second is an algorithmic code-generation problem from MBPP, representing a case where the controller must navigate genuine uncertainty before the solution structure becomes clear.

\begin{table*}[t]
\vspace{-0.5em}
\centering
\footnotesize
\def\arraystretch{1.12}
\setlength{\tabcolsep}{4.0pt}
\begin{tabular}{p{0.10\textwidth}p{0.25\textwidth}ccp{0.36\textwidth}}
\toprule
Stage & Local observation & Temperature & Budget & Condensed generated segment \\
\midrule
$t_0$ Think & New problem; math prior, no evidence yet about difficulty. & $0.7$ & $1024$ & Identify the task as rectangular-to-polar conversion and recall $r=\sqrt{x^2+y^2}$. \\
$t_1$ Think & Formula matched; entropy drops after substituting $x=0,y=3$. & $0.3$ & $256$ & Compute $r=\sqrt{0^2+3^2}=3$. Since the point lies on the positive $y$-axis, $\theta=\pi/2$. \\
$t_2$ Think & Both required fields are determined; additional reasoning is unnecessary. & $0.3$ & $0$ & Close the reasoning phase after checking that $\theta=\pi/2$ satisfies $0\leq\theta<2\pi$. \\
$t_3$ Answer & Reasoning has ended; budget control is frozen, but answer formatting remains active. & $0$ & $\varnothing_b$ & Output $\boxed{\left(3,\frac{\pi}{2}\right)}$ in the requested tuple format. \\
\bottomrule
\end{tabular}
\caption{Condensed controlled trajectory for the MATH-500 polar-coordinate example on \texttt{Qwen3-8B}. This example contains three control updates.}
\label{tab:case_math}
\end{table*}

\begin{table*}[t]
\centering
\footnotesize
\def\arraystretch{1.12}
\setlength{\tabcolsep}{4.0pt}
\begin{tabular}{p{0.10\textwidth}p{0.25\textwidth}ccp{0.36\textwidth}}
\toprule
Stage & Local observation & Temperature & Budget & Condensed generated segment \\
\midrule
$t_0$ Think & Code task detected; prompt underspecifies whether pairs are pre-sorted. & $0.7$ & $1024$ & Parse the required relation: pair $j$ can precede pair $i$ when $a_i>b_j$. Consider DP over pair positions. \\
$t_1$ Think & Competing strategies appear: greedy sorting is tempting but not guaranteed by the prompt. & $1.0$ & $4096$ & Compare solution templates and select the safer recurrence: \texttt{mcl[i]} is the best chain ending at \texttt{i}. \\
$t_2$ Think & Recurrence is fixed; implementation should be stable and syntax-sensitive. & $0.3$ & $1024$ & Generate the nested-loop function, initialize all chain lengths to $1$, and update \texttt{mcl[i] = mcl[j] + 1}. \\
$t_3$ Think & Code structure and tests are resolved; remaining work is submission formatting. & $0.3$ & $0$ & End the reasoning phase after preserving the \texttt{Pair.a}/\texttt{Pair.b} interface and the required function signature. \\
$t_4$ Answer & Reasoning budget is frozen; sampling controls deterministic code-body emission. & $0.3$ & $\varnothing_b$ & Emit the \texttt{Pair} class and the loop structure, keeping the generated code syntactically complete. \\
$t_5$ Answer & Code body is complete; only final submission closure remains. & $0$ & $\varnothing_b$ & Return \texttt{max(mcl)} after the loops and stop after the submitted program. \\
\bottomrule
\end{tabular}
\caption{Condensed controlled trajectory for the MBPP longest-chain code example on \texttt{Qwen3-8B}. The trajectory contains five control updates.}
\label{tab:case_code}
\vspace{-0.8em}
\end{table*}

\subsection{Example of Early Consolidation }
\label{app:case_math}

 The instance \texttt{test/precalculus/807.json} from MATH-500 asks: ``Convert the point $(0,3)$ in rectangular coordinates to polar coordinates. Enter your answer in the form $(r,\theta)$, where $r>0$ and $0\leq\theta<2\pi$.'' The correct answer is $\left(3,\frac{\pi}{2}\right)$.

This problem has a simple structure: once the controller observes that the prompt only requires a direct coordinate conversion, the useful reasoning path is short and deterministic. A fixed large-budget setting can still solve the problem, but it tends to spend unnecessary tokens restating the polar-coordinate formula and checking quadrants. The control process of \textbf{\texttt{AutoCRAT}} is demonstrated in Table \ref{tab:case_math}. The case illustrates why dynamic budget reduction matters even on easy examples. The first boundary keeps enough flexibility to recognize the task type, while the later boundaries become deterministic and short once the local state is resolved. This prevents overthinking without relying on a globally small budget that would be detrimental for harder math problems.

\subsection{Example of Exploratory-Consolidation}
\label{app:case_code}

The instance \texttt{task\_id=601} from MBPP asks the model to ``write a function to find the longest chain which can be formed from the given set of pairs.'' A valid solution defines a pair $(a,b)$ and returns the maximum number of pairs that can be chained so that the next pair's first element is larger than the previous pair's second element. The representative tests include chains such as
\texttt{[(5,24),(15,25),(27,40),(50,60)] -> 3}.

In this example, the model typically infers the algorithmic invariant, handles ordering assumptions, and produces executable Python. A low-budget deterministic rollout often jumps directly to code and risks missing the dynamic-programming recurrence, while an unconstrained high-budget rollout may spend many tokens discussing alternative greedy and sorting variants. As shown in Table \ref{tab:case_code}, \textbf{\texttt{AutoCRAT}} allocates more compute during algorithm selection, then reduces stochasticity during implementation. After reasoning ends, the budget action is frozen, but the temperature remains controlled across two answer-stage boundaries. This trajectory highlights the interaction between the two controlled axes. Furthermore, the answer phase demonstrates that temperature control remains meaningful even after budget actions are frozen, as deterministic emission is preferred for code body but a small allowance is retained during the transition to final closure. Related control dynamics appear in other sequential settings \cite{liu2026palm,yan2026openskillopenworldselfevolutionllm}.

\subsection{Failure Case: Premature Commitment}
\label{app:case_premature}

In MATH-500 case \texttt{test/algebra/2193.json} on \texttt{Qwen3-8B}, the model makes an early algebraic error while isolating the square-root term but continues along a locally coherent derivation. At the following semantic boundary, the prefix exhibits low entropy and high top-1 probability. The controller consequently selects a lower temperature and a smaller remaining budget. Subsequent decoding remains concentrated around the incorrect derivation, leaving insufficient room for a later self-check. This case illustrates that decoder confidence reflects local predictive concentration rather than correctness, and the controller can mistake a wrong but fluent intermediate state for convergence.

\subsection{Failure Case: Unproductive Continuation}
\label{app:case_unproductive}

In HumanEval case \texttt{has\_close\_elements} on \texttt{Qwen3-8B}, the model identifies the appropriate pairwise-distance scan but continues reconsidering equivalent nested-loop and pair-combination implementations. The remaining surface-form uncertainty keeps the decoder-side signals elevated, and the controller retains a large budget. The trajectory spends additional tokens on repeated planning without changing the resulting implementation. This case shows that token-level uncertainty can persist after the decision-relevant reasoning is already complete, causing compute over-allocation.

\section{Multi-Seed Stability}
\label{app:seeds}

We evaluate three controller-training seeds (42, 123, 2026) on \texttt{Qwen3-4B}. Seed 42 corresponds to the reported run in the main experiments. Table~\ref{tab:seeds} reports per-benchmark accuracy and average tokens. Every seed improves both accuracy and token efficiency over \textsc{Recommend} on all six benchmarks.

\begin{table*}[t]
\centering
\scriptsize
\def\arraystretch{1.05}
\setlength{\tabcolsep}{2.5pt}
\resizebox{0.99\textwidth}{!}{%
\begin{tabular}{l*{6}{cc}}
\toprule
\multirow{2}{*}{\scriptsize\bf Setting} & \multicolumn{2}{c}{\scriptsize\bf GSM8K} & \multicolumn{2}{c}{\scriptsize\bf MATH-500} & \multicolumn{2}{c}{\scriptsize\bf ARC-C} & \multicolumn{2}{c}{\scriptsize\bf GPQA-D} & \multicolumn{2}{c}{\scriptsize\bf HumanEval} & \multicolumn{2}{c}{\scriptsize\bf MBPP} \\
\cmidrule(r){2-3} \cmidrule(r){4-5} \cmidrule(r){6-7} \cmidrule(r){8-9} \cmidrule(r){10-11} \cmidrule(r){12-13}
& \scriptsize\bf Acc. & \scriptsize\bf Tok. & \scriptsize\bf Acc. & \scriptsize\bf Tok. & \scriptsize\bf Acc. & \scriptsize\bf Tok. & \scriptsize\bf Acc. & \scriptsize\bf Tok. & \scriptsize\bf Acc. & \scriptsize\bf Tok. & \scriptsize\bf Acc. & \scriptsize\bf Tok. \\
\midrule
\textsc{Recommend} & 92.0 & 2189 & 94.5 & 2765 & 68.4 & 1042 & 52.5 & 1276 & 67.2 & 2665 & 52.5 & 1708 \\
\midrule
Seed 42 & 93.1 & 729 & 95.8 & 1333 & 70.2 & 382 & 53.8 & 1220 & 71.0 & 1138 & 55.4 & 1091 \\
Seed 123 & 92.8 & 751 & 95.5 & 1367 & 69.7 & 395 & 53.2 & 1261 & 70.2 & 1176 & 54.5 & 1126 \\
Seed 2026 & 93.6 & 716 & 96.3 & 1306 & 70.9 & 374 & 55.7 & 1191 & 72.5 & 1107 & 56.3 & 1060 \\
\midrule
Mean$\pm$std & 93.2$\pm$0.4 & 732$\pm$18 & 95.9$\pm$0.4 & 1335$\pm$31 & 70.3$\pm$0.6 & 384$\pm$11 & 54.2$\pm$1.3 & 1224$\pm$35 & 71.2$\pm$1.2 & 1140$\pm$35 & 55.4$\pm$0.9 & 1092$\pm$33 \\
\bottomrule
\end{tabular}
}
\caption{Multi-seed stability on \texttt{Qwen3-4B}. Seed 42 is the reported run.}
\label{tab:seeds}
\end{table*}

\section{Practical Efficiency and Controller Overhead}
\label{app:latency}

We profile \texttt{Qwen3-4B} batch-one inference on a single NVIDIA A100-80GB, averaging three runs after warm-up under identical prompts and decoding infrastructure. Controller overhead accumulates boundary detection, feature construction, and control-head inference time. Results are in Table~\ref{tab:latency}.

\begin{table}[H]
\centering
\scriptsize
\def\arraystretch{1.05}
\setlength{\tabcolsep}{2.5pt}
\resizebox{0.99\columnwidth}{!}{%
\begin{tabular}{llccc}
\toprule
\textbf{Benchmark} & \textbf{Method} & \textbf{Throughput (tok/s)} & \textbf{Ctrl. overhead (s/q)} & \textbf{Wall-clock (s/q)} \\
\midrule
\multirow{2}{*}{GSM8K} & \textsc{Recommend} & 91.3 & -- & 24.0 \\
& \textbf{\texttt{AutoCRAT}} & 74.6 & 1.43 & 9.8 \\
\midrule
\multirow{2}{*}{ARC-C} & \textsc{Recommend} & 94.1 & -- & 11.1 \\
& \textbf{\texttt{AutoCRAT}} & 78.7 & 0.67 & 4.9 \\
\midrule
\multirow{2}{*}{HumanEval} & \textsc{Recommend} & 85.7 & -- & 31.1 \\
& \textbf{\texttt{AutoCRAT}} & 66.9 & 2.91 & 17.0 \\
\bottomrule
\end{tabular}
}
\caption{Controller overhead and end-to-end wall-clock time on \texttt{Qwen3-4B} (A100-80GB, batch one). Throughput decreases by 16--22\% due to controller overhead, but shorter generations reduce wall-clock time by 45--59\%.}
\label{tab:latency}
\end{table}

\section{Sensitivity to Reward-Cost Coefficient}
\label{app:lambda}

We vary the reward-cost coefficient $\lambda \in \{0.1, 0.2, 0.4\}$ on \texttt{Qwen3-4B}. All other hyperparameters are fixed; only the supervision targets and controller weights change. Results are in Table~\ref{tab:lambda}.

\begin{table}[H]
\centering
\scriptsize
\def\arraystretch{1.05}
\setlength{\tabcolsep}{3.0pt}
\resizebox{0.99\columnwidth}{!}{%
\begin{tabular}{c*{4}{cc}}
\toprule
\multirow{2}{*}{$\lambda$} & \multicolumn{2}{c}{\scriptsize\bf GSM8K} & \multicolumn{2}{c}{\scriptsize\bf ARC-C} & \multicolumn{2}{c}{\scriptsize\bf HumanEval} & \multicolumn{2}{c}{\scriptsize\bf Avg.} \\
\cmidrule(r){2-3} \cmidrule(r){4-5} \cmidrule(r){6-7} \cmidrule(r){8-9}
& \scriptsize\bf Acc. & \scriptsize\bf Tok. & \scriptsize\bf Acc. & \scriptsize\bf Tok. & \scriptsize\bf Acc. & \scriptsize\bf Tok. & \scriptsize\bf Acc. & \scriptsize\bf Tok. \\
\midrule
0.1 & 93.2 & 1061 & 70.2 & 493 & 71.8 & 1468 & 78.4 & 1007 \\
0.2 & 93.1 & 729 & 70.2 & 382 & 71.0 & 1138 & 78.1 & 750 \\
0.4 & 92.5 & 521 & 69.1 & 277 & 69.5 & 767 & 77.0 & 522 \\
\bottomrule
\end{tabular}
}
\caption{Sensitivity to the reward-cost coefficient $\lambda$ on \texttt{Qwen3-4B}. $\lambda{=}0.2$ (reported setting) provides a balanced operating point; $\lambda{=}0.4$ reduces tokens by 30\% at a larger accuracy cost.}
\label{tab:lambda}
\end{table}

\end{document}